\documentclass{article} % For LaTeX2e
\usepackage[accepted]{icml2017}
\usepackage{url}

\usepackage[utf8]{inputenc}

\usepackage{times}

\usepackage{booktabs}

\usepackage[subtle,tracking=normal]{savetrees}
\usepackage{subfig}

\usepackage{listings}
\usepackage{lstlinebgrd}
\usepackage{amsmath}
\usepackage{amssymb}
\usepackage{mathtools}
\usepackage{ifthen}
\usepackage{hyperref}       % hyperlinks
\hypersetup{
 unicode=false,          % non-Latin characters in Acrobat’s bookmarks
 pdftoolbar=true,        % show Acrobat’s toolbar?
 pdfmenubar=true,        % show Acrobat’s menu?
 pdffitwindow=false,     % window fit to page when opened
 pdfstartview={FitH},    % fits the width of the page to the window
 pdfnewwindow=true,      % links in new window
 colorlinks=true,        % false: boxed links; true: colored links
 linkcolor=black,        % color of internal links (change box color with linkbordercolor)
 citecolor=blue!50!black,    % color of links to bibliography
 filecolor=black,        % color of file links
 urlcolor=black          % color of external links
}
\usepackage{url}            % simple URL typesetting
\usepackage{graphicx}
\usepackage{subfig}
\usepackage{booktabs}
\usepackage{tikz}

\usepackage{enumitem} % for controlling enumerate indent

\definecolor{nice-red}{HTML}{E41A1C}
\definecolor{nice-orange}{HTML}{FF7F00}
\definecolor{nice-yellow}{HTML}{FFC020}
\definecolor{nice-green}{HTML}{4DAF4A}
\definecolor{nice-blue}{HTML}{377EB8}
\definecolor{nice-purple}{HTML}{984EA3}

\newboolean{blackandwhite}
\setboolean{blackandwhite}{false}
\ifthenelse{\boolean{blackandwhite}}{
  \colorlet{dark-blue}{black}
  \colorlet{dark-purple}{black}
  \colorlet{dark-red}{black}
  \colorlet{todo}{black}
  \colorlet{todoref}{black}
}{
  \colorlet{dark-blue}{blue!50!black}
  \colorlet{dark-purple}{purple!50!black}
  \colorlet{dark-red}{red!50!black}
  \colorlet{todo}{red!85!black}
  \colorlet{todoref}{purple!70!black}
}

\renewenvironment{quote}
  {\small\list{}{\rightmargin=0.4cm \leftmargin=0.4cm}%
  \item\relax}
  {\endlist}

\newcommand{\fnam}{$\partial$4}

\newboolean{draftmode}
\setboolean{draftmode}{false}

\ifthenelse{\boolean{draftmode}}{
  
  \newcommand{\todoc}[1]{\textcolor{todo}{\textbf{TODO} #1}}
  
  \newcommand{\torefc}[1]{\textcolor{todoref}{\textbf{REF} #1}}
  \newcommand{\tim}[1]{{\color{nice-blue}TR: #1}}
  \newcommand{\Seb}[1]{{\color{nice-red}SR: #1}}
  \newcommand{\jason}[1]{{\color{nice-purple}JN: #1}}
  \newcommand{\mat}[1]{{\color{nice-green}MB: #1}}
  \newcommand{\tocite}[1]{{\color{nice-red}Cite: #1}}
  
}{
  
  \newcommand{\todoc}[1]{}
  
  \newcommand{\torefc}[1]{}
  \newcommand{\tim}[1]{}
  \newcommand{\Seb}[1]{}
  \newcommand{\mat}[1]{}
  \newcommand{\jason}[1]{}
  \newcommand{\tocite}[1]{}
  
}

\newcommand{\ie}{{\emph{i.e.}}}

\newcommand{\stacks}{\mathcal{M}}
\newcommand{\buffer}{\mathbf{M}}
\newcommand{\pointer}{\mathbf{a}}
\newcommand{\x}{\mathbf{x}}

\newcommand{\dstack}{\mathcal{D}}
\newcommand{\rstack}{\mathcal{R}}

\newcommand{\pc}{\mathbf{c}}

\newcommand{\dstackb}{\mathbf{D}}
\newcommand{\rstackb}{\mathbf{R}}
\newcommand{\heapb}{\mathbf{H}}

\newcommand{\dstackt}{\mathbf{d}}
\newcommand{\dstackn}{\mathbf{d}^{-1}}
\newcommand{\rstackt}{\mathbf{r}}

\newcommand{\vstate}{\mathbf{S}}

\newcommand{\word}{\mathbf{w}}

\newcommand{\vS}{\mathcal{S}}

\let\othelstnumber=\thelstnumber
\def\createlinenumber#1#2{
    \edef\thelstnumber{%
        \unexpanded{%
            \ifnum#1=\value{lstnumber}\relax
              #2%
            \else}%
        \expandafter\unexpanded\expandafter{\thelstnumber\othelstnumber\fi}%
    }
    \ifx\othelstnumber=\relax\else
      \let\othelstnumber\relax
    \fi
}

\definecolor{mygreen}{RGB}{201, 255, 202}
\definecolor{myyellow}{RGB}{255, 254, 201}
\definecolor{myblue}{RGB}{201, 223, 255}

\makeatother

\lstdefinelanguage{Forth}{
    morekeywords={%
        BEGIN, WHILE, REPEAT, DO, LOOP, ';', IF, THEN, ELSE, observe, manipulate, choose, permute, ':'},%
    sensitive,%
    morecomment=[l]\\,%
    morecomment=[s]{(}{)},%
}[keywords,comments]%

\RequirePackage{textcomp}
\newcommand{\code}[1]{\texttt{#1}}
\newcommand{\stack}[1]{[{\color{blue!50!black}#1}]}

\icmltitlerunning{Programming with a Differentiable Forth Interpreter}

\icmltitlerunning{Programming with a Differentiable Forth Interpreter}

\begin{document}

%\maketitle
\twocolumn[\icmltitle{Programming with a Differentiable Forth Interpreter}
\icmlsetsymbol{equal}{*}

\begin{icmlauthorlist}
\icmlauthor{Matko Bošnjak}{ucl}
\icmlauthor{Tim Rocktäschel}{ox}
\icmlauthor{Jason Naradowsky}{cam}
\icmlauthor{Sebastian Riedel}{ucl}
\end{icmlauthorlist}

\icmlaffiliation{ucl}{Department of Computer Science, University College London, London, UK}
\icmlaffiliation{cam}{Department of Theoretical and Applied Linguistics, University of Cambridge, Cambridge, UK}
\icmlaffiliation{ox}{Department of Computer Science, University of Oxford, Oxford, UK}

\icmlcorrespondingauthor{Matko Bošnjak}{m.bosnjak@cs.ucl.ac.uk}
% \icmlcorrespondingauthor{Tim Rocktäschel}{t.rocktaschel@cs.ucl.ac.uk}
% \icmlcorrespondingauthor{Jason Naradowsky}{j.narad@cs.ucl.ac.uk}
% \icmlcorrespondingauthor{Sebastian Riedel}{s.riedel@cs.ucl.ac.uk}

\icmlkeywords{boring formatting information, machine learning, ICML}

\vskip 0.3in
]



\printAffiliationsAndNotice{}  % leave blank if no need to mention equal contribution
%\footnotetext{hi}

\begin{abstract}
Given that in practice training data is scarce for all but a small set of problems, a core question is how to incorporate prior knowledge into a model.
In this paper, we consider the case of prior \emph{procedural} knowledge for neural networks, such as knowing how a program should traverse a sequence, but not what local actions should be performed at each step.
To this end, we present an end-to-end differentiable interpreter for the programming language Forth which enables programmers to write program sketches with slots that can be filled with behaviour trained from program input-output data. 
We can optimise this behaviour directly through gradient descent techniques on user-specified objectives, and also integrate the program into any larger neural computation graph. 
We show empirically that our interpreter is able to effectively leverage different levels of prior program structure and learn complex behaviours such as sequence sorting and addition.
When connected to outputs of an LSTM and trained jointly, our interpreter achieves state-of-the-art accuracy for end-to-end reasoning about quantities expressed in natural language stories.

% SR: leaving this out for now. We should make these points in the intro, but in the abstract it's so prominent that I'd look for experiments... Our long-term goal is to exploit this knowledge to rapidly learn more complex computation such as natural language parsing or knowledge base population. 
\end{abstract}

\section{Introduction}

A central goal of Artificial Intelligence is the creation of machines that learn as effectively from human instruction as they do from data.
A recent and important step towards this goal is the invention of neural architectures that learn to perform algorithms akin to traditional computers, using primitives such as memory access and stack manipulation~\citep{graves2014neural, joulin2015inferring,grefenstette2015learning,kaiser2015neural, kurach2015neural, graves2016hybrid}.
These architectures can be trained through standard gradient descent methods, and enable machines to learn complex behaviour from input-output pairs or program traces.
In this context, the role of the human programmer is often limited to providing training data.
However, training data is a scarce resource for many tasks.
In these cases, the programmer may have \emph{partial} procedural background knowledge: one may know the rough structure of the program, or how to implement several subroutines that are likely necessary to solve the task.
For example, in programming by demonstration \citep{lau2001learning} or query language programming~\citep{neelakantan2015neural} a user establishes a larger set of conditions on the data, and the model needs to set out the details.
In all these scenarios, the question then becomes how to exploit various types of prior knowledge when learning algorithms.

To address the above question we present an approach that enables programmers to inject their procedural background knowledge into a neural network. In this approach, the programmer specifies a program \emph{sketch}~\citep{solar2005programming} in a traditional programming language. This sketch defines one part of the neural network behaviour. The other part is learned using training data. The core insight that enables this approach is the fact that most programming languages can be formulated in terms of an abstract machine that 
%programming languages can usually be formulated in terms of an abstract machine that 
executes the commands of the language. We implement these machines as neural networks, constraining parts of the networks to follow the sketched behaviour. The resulting neural programs are consistent with our prior knowledge and optimised with respect to the training data. 

In this paper, we focus on the programming language Forth~\citep{brodie1980starting}, a simple yet powerful stack-based language that facilitates factoring and abstraction. Underlying Forth's semantics is a simple abstract machine. We introduce \fnam{}, an implementation of this machine that is differentiable with respect to the transition it executes at each time step, as well as distributed input representations. Sketches that users write define underspecified behaviour which can then be trained with backpropagation.

For two neural programming tasks introduced in previous work~\citep{reed2015neural} we present Forth sketches that capture different degrees of prior knowledge. For example, we define only the general recursive structure of a sorting problem. We show that given only input-output pairs, \fnam{} can learn to fill the sketch and generalise well to problems of unseen size. 
In addition, we apply \fnam{} to the task of solving word algebra problems.
We show that when provided with basic algorithmic scaffolding and trained jointly with an upstream LSTM~\cite{hochreiter1997long}, \fnam{} is able to learn to read natural language narratives, extract important numerical quantities, and reason with these, ultimately answering corresponding mathematical questions without the need for explicit intermediate representations used in previous work.

The contributions of our work are as follows: i) We present a neural implementation of a dual stack machine underlying Forth, ii) we introduce Forth sketches for programming with partial procedural background knowledge, iii) we apply Forth sketches as a procedural prior on learning algorithms from data, iv) we introduce program code optimisations based on symbolic execution that can speed up neural execution, and v) using Forth sketches we obtain state-of-the-art for end-to-end reasoning about quantities expressed in natural language narratives.

\section{The Forth Abstract Machine}

Forth is a simple Turing-complete stack-based programming language~\citep{ansforth1994, brodie1980starting}.
We chose Forth as the host language of our work because i) it is an established, general-purpose high-level language relatively close to machine code, ii) it promotes highly modular programs through use of branching, loops and function calls, thus bringing out a good balance between assembly and higher level languages, and importantly iii) its abstract machine is simple enough for a straightforward creation of its continuous approximation.
Forth's underlying abstract machine is represented by a state $S=(D,R,H,c)$, which contains two stacks: a data evaluation pushdown stack $D$ (\emph{data stack}) holds values for manipulation, and a return address pushdown stack $R$ (\emph{return stack}) assists with return pointers and subroutine calls.
These are accompanied by a \emph{heap} or random memory access buffer $H$, and a program counter $c$.

A Forth program $P$ is a sequence\footnote{Forth is a concatenative language.} of Forth \emph{words} (i.e. commands) $P=w_1 \ldots w_n$. %, that operate on a dual-stack Forth abstract machine. 
The role of a word varies, encompassing language keywords, primitives, and user-defined subroutines (e.g. \code{DROP} discards the top element of the data stack, or \code{DUP} duplicates the top element of the data stack).\footnote{In this work, we restrict ourselves to a subset of all Forth words, detailed in Appendix~\ref{appendix:forth-words}.} 
Each word $w_i$ defines a transition function between machine states $w_i:S \rightarrow S$.
Therefore, a program $P$ itself defines a transition function by simply applying the word at the current program counter to the current state.
Although usually considered as a part of the heap $H$, we consider Forth programs $P$ separately to ease the analysis.

An example of a Bubble sort algorithm implemented in Forth is shown in Listing~\ref{bubblesort} in everything except lines 3b-4c.
The execution starts from line 12 where literals are pushed on the data stack and the \code{SORT} is called. Line 10 executes the main loop over the sequence. Lines 2-7 denote the \code{BUBBLE} procedure -- comparison of top two stack numbers (line 3a), and the recursive call to itself (line 4a).
A detailed description of how this program is executed by the Forth abstract machine is provided in Appendix~\ref{appendix:bubblesort}.
Notice that while Forth provides common control structures such as looping and branching, these can always be reduced to low-level code that uses jumps and conditional jumps (using the words \code{BRANCH} and \code{BRANCH0}, respectively).
Likewise, we can think of sub-routine definitions as labelled code blocks, and their invocation amounts to jumping to the code block with the respective label.

\bgroup
\createlinenumber{3}{3a}
\createlinenumber{4}{4a}
\createlinenumber{5}{3b}
\createlinenumber{6}{4b}
\createlinenumber{7}{3c}
\createlinenumber{8}{4c}
\createlinenumber{9}{5}
\createlinenumber{10}{6}
\createlinenumber{11}{7}
\createlinenumber{12}{8}
\createlinenumber{13}{9}
\createlinenumber{14}{10}
\createlinenumber{15}{11}
\createlinenumber{16}{12}

\begin{figure}[t!]
%  \begin{minipage}{0.43\textwidth}
\begin{lstlisting}[
  mathescape,
  caption={Three code alternatives (white lines are common to all, coloured/lettered lines are alternative-specific): i) Bubble sort in Forth (a lines -- green), ii) 
  {\sc Permute} sketch (b lines -- blue), and iii) {\sc Compare} sketch (c lines -- yellow).},
  label=bubblesort,
  frame=single,
  basicstyle=\ttfamily\scriptsize,
  linebackgroundcolor={\ifnum\value{lstnumber}=3\color{mygreen}\else\ifnum\value{lstnumber}=4\color{mygreen}\else\ifnum\value{lstnumber}=5\color{myblue}\else\ifnum\value{lstnumber}=6\color{myblue}\else\ifnum\value{lstnumber}=7\color{myyellow}\else\ifnum\value{lstnumber}=8\color{myyellow}\fi\fi\fi\fi\fi\fi}
]
: BUBBLE ( a1 ... an n-1 -- one pass )
    DUP IF >R
        OVER OVER < IF SWAP THEN
        R> SWAP >R 1- BUBBLE R>
        { observe D0 D-1 -> permute D-1 D0 R0}
        1- BUBBLE R>
        { observe D0 D-1 -> choose NOP SWAP }
        R> SWAP >R 1- BUBBLE R>
    ELSE
        DROP
    THEN
;
: SORT ( a1 .. an n -- sorted  )   
  1- DUP 0 DO >R R@ BUBBLE R> LOOP DROP
;
2 4 2 7 4 SORT \ Example call
\end{lstlisting}  
\end{figure}
\egroup

%Notice that while Forth provides common control structures such as looping and branching, these can always be reduced to low-level code that uses jumps and conditional jumps (using the words \code{BRANCH} and \code{BRANCH0}, respectively). Likewise, we can think of sub-routine definitions as code blocks tagged with a label, and their invocation amounts to jumping to the tagged label.

%\fnam{}: A Differentiable Forth Abstract Machine
\section{\fnam{}: Differentiable Abstract Machine}

\begin{figure*}
\centering
\includegraphics[scale=0.8]{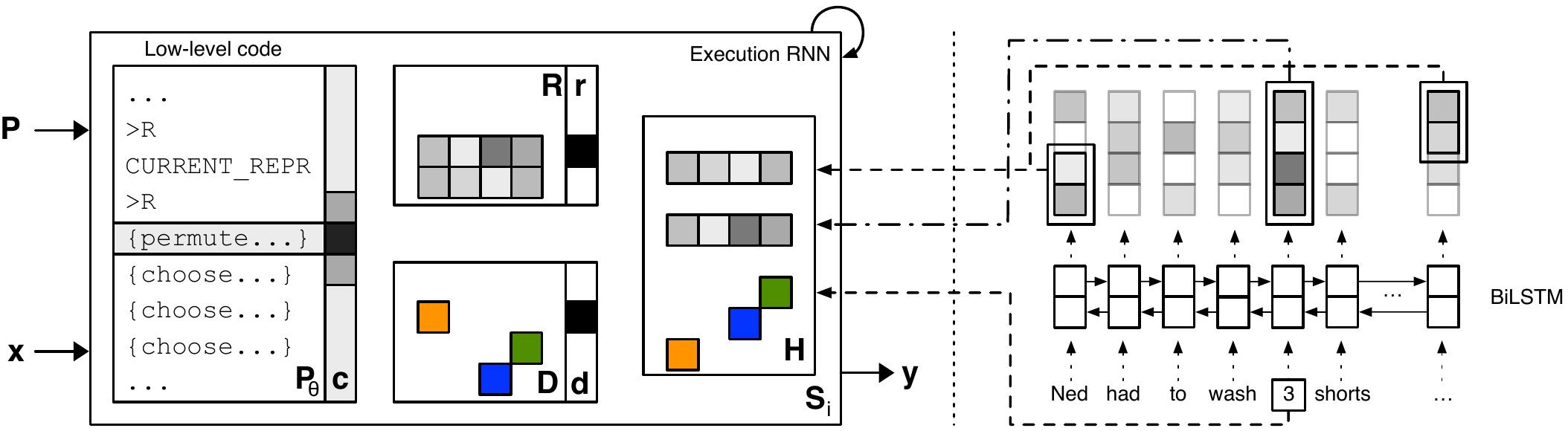}
\caption {\emph{Left:} Neural Forth Abstract Machine. A forth sketch $\mathbf{P}$ is translated to a low-level code $\mathbf{P}_{\theta}$, with slots \code{\{...\}} substituted by a parametrised neural networks. Slots are learnt from input-output examples $(\mathbf{x}, \mathbf{y})$ through the differentiable machine whose state $\mathbf{S}_{i}$ comprises the low-level code, program counter $\mathbf{c}$, data stack $\dstackb$ (with pointer $\mathbf{d}$), return stack $\rstackb$ (with pointer $\mathbf{r}$), and the heap $\heapb$. \emph{Right:} BiLSTM trained on Word Algebra Problems. Output vectors corresponding to a representation of the entire problem, as well as context representations of numbers and the numbers themselves are fed into $\heapb$ to solve tasks. The entire system is end-to-end differentiable.}
\label{model}
\end{figure*}

% \begin{figure}
% %  \end{minipage}\hspace{4.0em}
% %  \begin{minipage}{0.43\textwidth}
% \begin{lstlisting}[
%   mathescape,
%   caption=\code{BUBBLE} sketch with trainable permutation (trainable comparison in comments).,
%   label=bubblesketch,
%   frame=single,
%   basicstyle=\ttfamily\scriptsize
% ]
% : BUBBLE ( a1 ... an n-1 -- one pass )
%     DUP IF >R
%         { observe D0 D-1 -> permute D-1 D0 R0 }
%         1- BUBBLE R>
%         \ ** Alternative sketch **
%         \ { observe D0 D-1 -> choose NOP SWAP }
%         \ R> SWAP >R 1- BUBBLE R>
%     ELSE
%         DROP
%     THEN
% ;
% \end{lstlisting}  
% %  \end{minipage}
% \end{figure}

When a programmer writes a Forth program, they define a sequence of Forth words, \ie{}, a sequence of \emph{known} state transition functions. In other words, the programmer knows \emph{exactly} how computation should proceed. To accommodate for cases when the developer's procedural background knowledge is incomplete, we extend Forth to support the definition of a program \emph{sketch}. As is the case with Forth programs, sketches are sequences of transition functions. However, a sketch may contain transition functions whose behaviour is learned from data.

To learn the behaviour of transition functions within a program we would like the machine output to be differentiable with respect to these functions (and possibly representations of inputs to the program). This enables us to choose parameterised transition functions such as neural networks. 

To this end, we introduce \fnam{}, a TensorFlow~\citep{tensorflow2015-whitepaper} implementation of a differentiable abstract machine with continuous state representations, differentiable words and sketches. Program execution in \fnam{} is modelled by a recurrent neural network (RNN), parameterised by the transition functions at each time step.

%We then present a recurrent neural network (RNN) that models program execution on this machine, parametrised by the transition functions at each time step.

%of the first provide a continuous representation of the state of a Forth abstract machine. We then present a recurrent neural network (RNN) that models program execution on this machine, parametrised by the transition functions at each time step. 
%program index.\tim{first introduction "program index"}
%We refer to the model comprising the differentiable Forth abstract machine and its execution as \fnam{}.
%Lastly, we discuss optimizations based on symbolic execution and the interpolation of conditional branches.

\subsection{Machine State Encoding}

We map the symbolic machine state $S=(D,R,H,c)$ to a continuous representation $\vstate = (\dstack, \rstack, \heapb, \pc)$ into two differentiable stacks (with pointers), the data stack $\dstack=(\dstackb, \dstackt)$ and the return stack $\rstack=(\rstackb, \rstackt)$, a heap $\heapb$, and an attention vector $\pc$ indicating which word of the sketch $\mathbf{P}_{\theta}$ is being executed at the current time step.
Figure~\ref{model} depicts the machine together with its elements.
All three memory structures, the data stack, the return stack and the heap, are based on differentiable flat memory buffers $\buffer \in \{ \dstackb, \rstackb, \heapb \}$, where $\dstackb, \rstackb, \heapb \in \mathbb{R}^{l \times v}$, for a stack size $l$ and a value size $v$.
Each has a differentiable read operation
\begin{equation*}
    \text{read}_{\buffer}(\pointer) = \pointer^{T} \buffer
\end{equation*}
and write operation
\begin{equation*}
    \text{write}_{\buffer}(\x, \pointer): \buffer \leftarrow \buffer - 
    (\pointer \mathbf{1}^{T}) \odot \buffer + \x \pointer^{T}
\end{equation*}
akin to the Neural Turing Machine (NTM) memory~\citep{graves2014neural}, where $\odot$ is the element-wise multiplication, and $\pointer$ is the address pointer.\footnote{The equal widths of $\heapb$ and $\dstackb$ allow us to directly move vector representations of values between the heap and the stack.}
In addition to the memory buffers $\dstackb$ and $\rstackb$, the data stack and the return stack contain pointers to the current top-of-the-stack (TOS) element $\dstackt, \rstackt \in \mathbb{R}^l$, respectively. 
This allows us to implement pushing as writing a value $\mathbf{x}$ into $\buffer$ and incrementing the TOS pointer as:
\begin{equation*}
    \text{push}_{\buffer}(\x):
        \text{write}_{\buffer}(\x, \mathbf{p})
        \tag{side-effect: $\mathbf{p} \leftarrow \text{inc}(\mathbf{p})$ }
\end{equation*}
where $\mathbf{p} \in \{\dstackt, \rstackt\}$, $\text{inc}(\mathbf{p}) = \mathbf{p}^{T} \mathbf{R^1+} $, $\text{dec}(\mathbf{p}) = \mathbf{p}^{T} \mathbf{R^-} $, and $\mathbf{R^1+}$ and $\mathbf{R^1-}$ are increment and decrement matrices (left and right circular shift matrices).

Popping is realized by multiplying the TOS pointer and the memory buffer, and decreasing the TOS pointer:
% POP an element from $\stacks$ & 
\begin{equation*}
    \text{pop}_{\buffer}(\ ) = \text{read}_{\buffer}(\mathbf{p})
    \tag{side-effect: $\mathbf{p} \leftarrow \text{dec}(\mathbf{p})$}
\end{equation*}

% Since the data $D$ and return stack $R$ are both represented using the same mechanism, we will only describe $D$. 
% Its continuous representation $\dstack$ is a tuple $(\dstackb, \dstackt)$ where $\dstackb \in \mathbb{R}^{l \times v}$ is a matrix which serves as a buffer of length $l$ and value width $v$.  $\dstackt \in \mathbb{R}^l$ is a pointer to the current top element in the stack as contained in the buffer.

% To access the top element of the stack we use the product $\dstackt \dstackb$. Pushing and popping is implemented straightforwardly with well defined differentiable reading and writing procedures, similarly to the Neural Turing Machine memory~\citep{graves2014neural} by writing into the buffer matrix and incrementing and decrementing $\dstackt$, respectively:

Finally, the program counter $\pc \in \mathbb{R}^p$ is a vector that, when one-hot, points to a single word in a program of length $p$, and is equivalent to the $c$ vector of the symbolic state machine.\footnote{During training $\pc$ can become distributed and is considered as attention over the program code.} We use $\vS$ to denote the space of all continuous representations $\vstate$.

\paragraph{Neural Forth Words}
It is straightforward to convert Forth words, defined as functions on discrete machine states, to functions operating on the continuous space $\vS$.
For example, consider the word \code{DUP}, which duplicates the top of the data stack.  A differentiable version of \code{DUP} first calculates the value $\mathbf{e}$ on the TOS address of $\dstackb$, as $\mathbf{e} = \dstackt^{T} \dstackb$.  It then shifts the stack pointer via $\dstackt \leftarrow \text{inc}(\dstackt)$, and writes $\mathbf{e}$ to $\dstackb$ using $\text{write}_{\dstackb}(\mathbf{e}, \dstackt)$. 
The complete description of implemented Forth Words and their differentiable counterparts can be found in Appendix~\ref{appendix:fnam-words}.

% It is easy to convert Forth words, defined as functions on discrete machine states, to functions operating on the continuous space $\vS$. For example, for \code{DROP} we simply multiply the stack top pointer $\dstackt$ with a circular permutation matrix $\buffer^{-}$ that circularly shifts the vector.
% Likewise, for \code{DUP} we calculate the top stack element $e = \dstackt \dstackb$, 
% shift the stack pointer $\dstackt$ by multiplying it with $\buffer^{+}$, and then differentiably writing $e$ into $\dstackb$. at the new top pointer position. For exact expressions on all FNAM commands see Appendix TODO.

\subsection{Forth Sketches}
\label{sketch}

We define a Forth sketch $\mathbf{P_{\theta}}$ as a sequence of continuous transition functions $\mathbf{P}=\word_1 \ldots \word_n$. 
Here, $\word_i \in \vS\rightarrow\vS$ either corresponds to a neural Forth word or a trainable transition function (neural networks in our case). We will call these trainable functions \emph{slots}, as they correspond to underspecified ``slots'' in the program code that need to be filled by learned behaviour. 

% Good sentence - wrong place?
%Our goal is to define a neural version of the Forth abstract machine that can execute such programs and guarantees that both symbolic execution of the neural program and generic execution of a standard Forth program result in equivalent final states.

%when executed symbolically, leads to a final state equivalent to the execution of a standard Forth program.
%guarantees that the execution of a standard Forth program leads to a final state equivalent to the final state of the corresponding symbolic execution.

% \subsection{Forth Sketches}\label{sketch}

We allow users to define a slot $\word$ by specifying a pair of a state encoder $\word_\text{enc}$ and a decoder $\word_\text{dec}$. The encoder produces a latent representation $\mathbf{h}$ of the current machine state using a multi-layer perceptron, and the decoder consumes this representation to produce the next machine state.
We hence have $\word = \word_\text{dec} \circ \word_\text{enc}$. To use slots within Forth program code, we introduce a notation that reflects this decomposition. In particular, slots are defined by the syntax \code{\{ encoder -> decoder \}} where \code{encoder} and \code{decoder} are specifications of the corresponding slot parts as described below. 

\paragraph{Encoders} We provide the following options for encoders:
\begin{description}[itemsep=0pt,parsep=0pt,topsep=0pt]
	\item[\code{static}] produces a static representation, independent of the actual machine state. 
	\item[\code{observe}] $\,e_1 \ldots e_m$: concatenates the elements $e_1 \ldots e_m$ of the machine state. An element can be a stack item \code{Di} at relative index $i$, a return stack item \code{Ri}, etc.
	\item[\code{linear N, sigmoid, tanh}] represent chained trans-formations, which enable the multilayer perceptron architecture. Linear N projects to N dimensions, and sigmoid and tanh apply same-named functions elementwise.
\end{description}  

\paragraph{Decoders}Users can specify the following decoders:
\begin{description}[itemsep=0pt,parsep=0pt,topsep=0pt]
	\item[\code{choose}] $\,w_1 \ldots w_m$: chooses from the Forth words $w_1 \ldots w_m$. Takes an input vector $\mathbf{h}$ of length $m$ to produce a weighted combination of machine states $\sum_i^m h_i \mathbf{w_i}(\vstate)$.
	\item[\code{manipulate}] $\,e_1 \ldots e_m$: directly manipulates the machine state elements $e_1 \ldots e_m$ by writing the appropriately reshaped and softmaxed output of the encoder over the machine state elements with $\text{write}_{\buffer}$.
	\item[\code{permute}] $\,e_1 \ldots e_m$: permutes the machine state elements $e_1 \ldots e_m$ via a linear combination of $m!$ state vectors.
\end{description}

% To give an example, consider the slot \code{\{ observe D0 D-1 -> choose 1+ 1- \}}. This slot uses the top (\code{D0}) and second from top (\code{D-1}) elements as state representation to determine whether to execute \code{1+} or \code{1-}. We can see this slot notation in the sketch of listing TODO.   

% \Seb{We should have some interesting "real world" examples of sketches here...}
% \tim{I like the differentiable BLEU score example}

\subsection{The Execution RNN}

We model execution using an RNN which produces a state $\vstate_{n+1}$ conditioned on a previous state $\vstate_n$. 
It does so by first passing the current state to each function $\word_i$ in the program, and then weighing each of the produced next states by the component of the program counter vector $\pc_i$ that corresponds to program index $i$, effectively using $\pc$ as an attention vector over code. Formally we have:
\begin{equation*}
    \vstate_{n+1} = \text{RNN}(\vstate_n, \mathbf{P}_{\theta}) = \sum_{i=1}^{|P|} \pc_i \word_i(\vstate_n)
% 	\vstate_{i+1} = \sum_{i} \pc_i \word_i(\vstate_i). 	
\end{equation*}
Clearly, this recursion, and its final state, are differentiable with respect to the program code $\mathbf{P}_{\theta}$, and its inputs.
Furthermore, for differentiable Forth programs the final state of this RNN will correspond to the final state of a symbolic execution (when no slots are present, and one-hot values are used).

\begin{figure}
\centering
\includegraphics[scale=0.467]{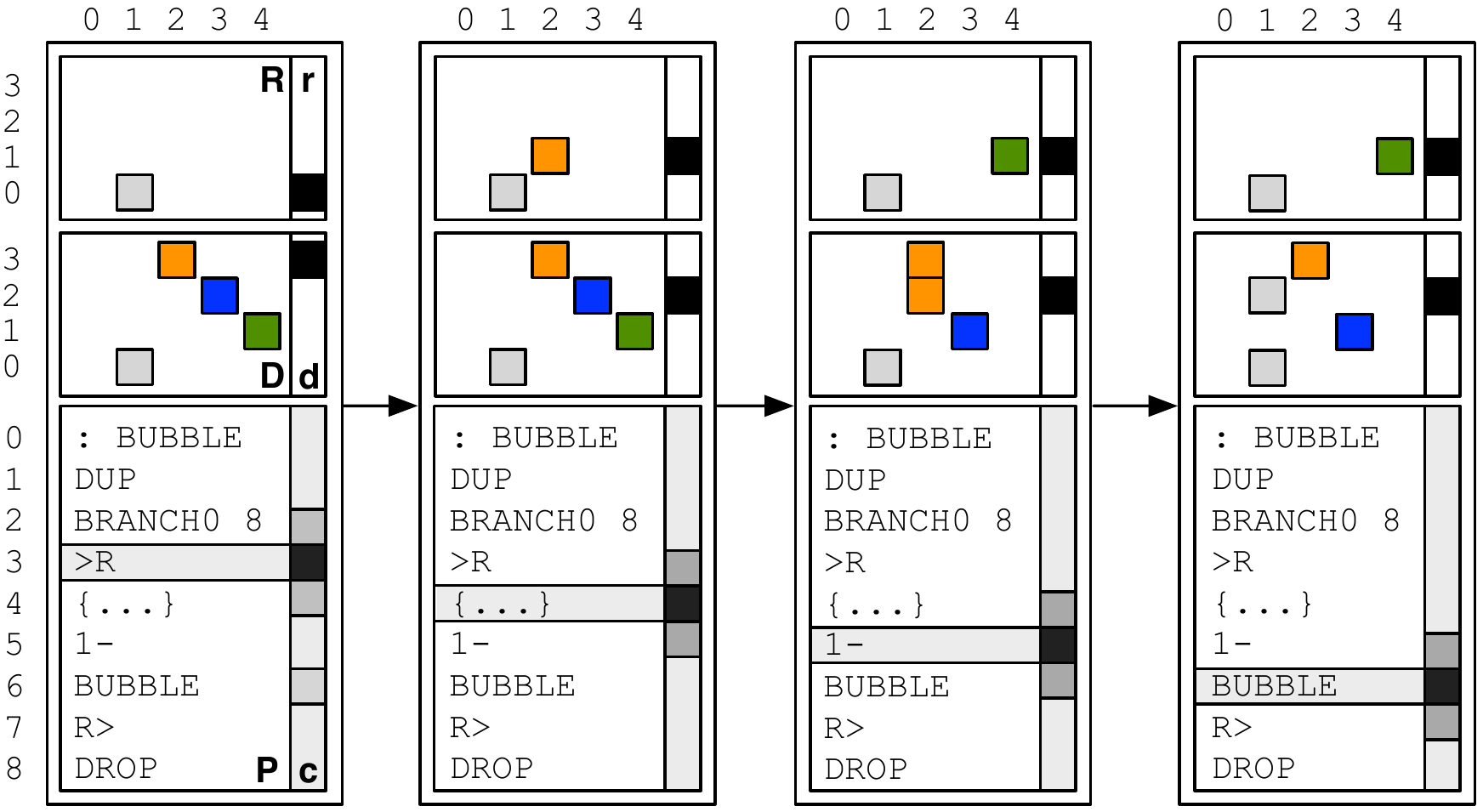}
\caption {
\fnam{} 
segment of the RNN execution of a Forth sketch in blue in Listing \ref{bubblesort}. The pointers ($\dstackt$, $\rstackt$) and values (rows of $\rstackb$ and $\dstackb$) are all in one-hot state (colours simply denote values observed, defined by the top scale), while the program counter maintains the uncertainty.  Subsequent states are discretised for clarity. Here, the slot \code{\{...\}} has learned its optimal behaviour. 
%Note that the heap $\heapb$ is omitted, and the last step should have added the value 6 on $\rstackb$, but it is not shown as it is out of the domain.
} \label{rnn}
\end{figure}

\subsection{Program Code Optimisations}
\label{sec:optim}
The \fnam{} RNN requires one-time step per transition.
After each time step, the program counter is either incremented, decremented, explicitly set or popped from the stack.
In turn, a new machine state is calculated by executing all words in the program and then weighting the result states by the program counter. %activation of the program counter at the given word.
As this is expensive, it is advisable to avoid full RNN steps wherever possible.
% This parallel execution of all words is expensive, and it is therefore advisable to avoid full RNN steps wherever possible.
We use two strategies to avoid full RNN steps and significantly speed-up \fnam{}: symbolic execution and interpolation of if-branches.

\paragraph{Symbolic Execution} Whenever we have a sequence of Forth words that contains no branch entry or exit points, we can collapse this sequence into a single transition instead of naively interpreting words one-by-one.
We symbolically execute~\citep{King:1976:SEP:360248.360252} a sequence of Forth words to calculate a new machine state. We then use the difference between the new and the initial state to derive the transition function of the sequence.
For example, the sequence \code{R> SWAP >R} that swaps top elements of the data and the return stack yields the symbolic state $D=r_1 d_2 \ldots d_l$. and $R=d_1 r_2 \ldots r_l$. 
Comparing it to the initial state, we derive a single neural transition that only needs to swap the top elements of $\dstackb$ and $\rstackb$.

\paragraph{Interpolation of If-Branches} We cannot apply symbolic execution to code with branching points as the branching behaviour depends on the current machine state, and we cannot resolve it symbolically.
However, we can still collapse if-branches that involve no function calls or loops by executing both branches in parallel and weighing their output states by the value of the condition.
If the if-branch does contain function calls or loops, we simply fall back to execution of all words weighted by the program counter.

\subsection{Training}
Our training procedure assumes input-output pairs of machine start and end states $(\mathbf{x}_i,\mathbf{y}_i)$ only.
The output $\mathbf{y}_i$ defines a target memory $\mathbf{Y}^{D}_i$ and a target pointer $\mathbf{y}^{d}_i$ on the data stack $\dstackb$.
Additionally, we have a mask $\mathbf{K}_i$ that indicates which components of the stack should be included in the loss (e.g. we do not care about values above the stack depth). 
We use $\dstackb_T(\theta, \mathbf{x}_i)$ and $\dstackt_T(\theta, \mathbf{x}_i)$ to denote the final state of $\dstackb$ and $\dstackt$ after $T$ steps of execution RNN and using an initial state $\mathbf{x}_i$. 
We define the loss function as
% $L(\vstate_T(\mathbf{x}_i), \mathbf{y}_i, \mathbf{m}_i )$. 
\begin{equation*}
\begin{multlined}
\mathcal{L}(\theta) = \mathcal{H}(\mathbf{K}_i \odot \dstackb_T(\theta, \mathbf{x}_i), \mathbf{K}_i \odot \mathbf{Y}^{D}_i) \\
+ \mathcal{H}(\mathbf{K}_i \odot \dstackt_T(\theta, \mathbf{x}_i), \mathbf{K}_i \odot \mathbf{y}^{d}_i)
\end{multlined}
\end{equation*}
where $\mathcal{H}(\mathbf{x}, \mathbf{y})= -\mathbf{x} \log{\mathbf{y}}$ is the cross-entropy loss, and $\theta$ are parameters of slots in the program $P$.
We can use backpropagation and any variant of gradient descent to optimise this loss function.
Note that at this point it would be possible to include supervision of the intermediate states (trace-level), as done by the Neural Program Interpreter~\cite{reed2015neural}.

% Note that it is trivial to also provide supervision of the intermediate states (trace-level), as done by the Neural Program Interpreter~\cite{reed2015neural}.

%\subsection{Word Encoding and State Transitions}
%\Seb{This can contain the current encoding, Matko's or both.}
%
%In general:
%
%\begin{center}
%  \begin{tabular}{| l || c | r |}
%    \hline
%    OPCODE & ADDRESS & LITERAL \\
%    \hline
%  \end{tabular}
%\end{center}
%
%
%where ADDRESS AND LITERAL are one-hot vectors
%
%However, depending on the OPCODE, there are 2 representations we used:
%
%\subsubsection{Machine instruction -like representation}
%
%\begin{center}
%  \begin{tabular}{| c | c | c | c | c | c | c | c |}
%    \hline
%    SP MOVE & RP MOVE & PC MOVE & SP WRITE & RP WRITE & N WRITE & RAM WRITE \\
%    \hline
%  \end{tabular}
%\end{center}
%
%OPCODE contains multiple one-hot vector fields which denote different operations.
%
%there are specific combinations which constitute a command, but also other unwanted combinations
%
%\subsubsection{One-hot representation}
%
%OPCODE is a single one-hot vector
%each element denotes a single
%
%OPCODE is a length 16 vector
%
%first 9 commands are Forth words
%the rest are low-level commands

\section{Experiments}

We evaluate \fnam{} on three tasks.  Two of these are simple transduction tasks, sorting and addition as presented in \cite{reed2015neural}, with varying levels of program structure.  For each problem, we introduce two sketches.  

We also test \fnam{} on the more difficult task of answering word algebra problems. We show that not only can \fnam{} act as a standalone solver for such problems, bypassing the intermediary task of producing formula templates which must then be executed, but it can also outperform previous work when trained on the same data.

\subsection{Experimental Setup}

Specific to the transduction tasks, we discretise memory elements during testing.
This effectively allows the trained model to generalise to any sequence length if the correct sketch behaviour has been learned.
We also compare against a Seq2Seq~\citep{Sutskever:2014} baseline. Full details of the experimental setup can be found in Appendix~\ref{appendix:experimental-details}.

%To illustrate the generalization ability of this architecture, 

\subsection{Sorting}

Sorting sequences of digits is a hard task for RNNs, as they fail to generalise to sequences even marginally longer than the ones they have been trained on \citep{reed2015neural}.
We investigate several strong priors based on Bubble sort for this transduction task and present two \fnam{} sketches in Listing~\ref{bubblesort} that enable us to learn sorting from only a few hundred training examples (see Appendix~\ref{appendix:accuracy-examples} for more detail):

\begin{description}[leftmargin=0.5cm,itemsep=0pt,parsep=0pt]
  \item {\sc Permute}. A sketch specifying that the top two elements of the stack, and the top of the return stack must be permuted based on the values of the former (line 3b). Both the value comparison and the permutation behaviour must be learned. The core of this sketch is depicted in Listing~\ref{bubblesort} (b lines), and the sketch is explained in detail in Appendix~\ref{appendix:bubble-execution}.

%Listing \ref{bubblesketch} shows a sketch for \code{BUBBLE} where we only specify that based on the top two elements of the stack, these two elements and the top of the return stack need to be permuted (line 2).
%The exact behavior needs to be learned end-to-end from input-output examples.
%In the next line we further specify that the result should be put on the return stack and \code{BUBBLE} should be called on the the rest of the stack (line 3).

  \item {\sc Compare}. This sketch provides additional prior procedural knowledge to the model. In contrast to {\sc Permute}, only the comparison between the top two elements on the stack must be learned (line 3c). The core of this sketch is depicted in Listing \ref{bubblesort} (c lines).
%   (lines 5 and 6),
  
  %with the commented out sketch in . 
%Here we are providing more structure and leave less behavior open to be learned.
\end{description}

In both sketches, the outer loop can be specified in \fnam{} (Listing \ref{bubblesort}, line 10), which repeatedly calls a function \code{BUBBLE}.  
In doing so, it defines sufficient structure so that the behaviour of the network is invariant to the input sequence length. 
 
 %Concretely, there the \fnam{} model only has to learn to compare the two top elements on the stack from input-output examples.

%This is achieved by defining sufficient structure so that the network's behavior is invariant to the input sequence length.
%For instance, we can assume that for sorting a list of numbers we need to repeatedly apply a function to gradually shorter sub-lists.

%\subsubsection{Quantitative Evaluation}

\begin{table}
\centering
\caption{Accuracy (Hamming distance) of Permute and Compare sketches in comparison to a Seq2Seq baseline on the sorting problem.}
\label{table:sorter-results}
\resizebox{\columnwidth}{!}{
\begin{tabular}{l rrr rrr}
\toprule
 & \multicolumn{3}{c}{Test Length 8} & \multicolumn{3}{c}{Test Length: 64} \\
 \cmidrule(l){2-4}
 \cmidrule(l){5-7}
 Train Length: & 2 & 3 & 4 & 2 & 3 & 4 \\
 \midrule
 Seq2Seq & 26.2 & 29.2 & 39.1 & 13.3 & 13.6 & 15.9 \\
 \fnam{} Permute & 100.0 & 100.0 & 19.82 & 100.0 & 100.0 & 7.81 \\
 \fnam{} Compare & 100.0 & 100.0 & 49.22 & 100.0 & 100.0 & 20.65 \\
% Seq2Seq w/ Attention & \\
\bottomrule
\end{tabular}
}
\end{table}

\paragraph{Results on Bubble sort}

A quantitative comparison of our models on the Bubble sort task is provided in Table~\ref{table:sorter-results}.
For a given test sequence length, we vary the training set lengths to illustrate the model's ability to generalise to sequences longer than those it observed during training.
We find that {\fnam} quickly learns the correct sketch behaviour, and it is able to generalise perfectly to sort sequences of $64$ elements after observing only sequences of length two and three during training.
In comparison, the Seq2Seq baseline falters when attempting similar generalisations, and performs close to chance when tested on longer sequences.
Both {\fnam} sketches perform flawlessly when trained on short sequence lengths, but under-perform when trained on sequences of length 4 due to arising computational difficulties ({\sc Compare} sketch performs better due to more structure it imposes). We discuss this issue further in Section~\ref{sec:discussion}.

%presents the comparison of the {\sc Permute} and {\sc Compare} sketches, together with the Seq2Seq baseline.

%We show this in Figure~\ref{fig:res1}, on both train and test accuracy, using train sequences of length 3 and test sequences of length 8.  When prior knowledge is provided ({\sc Compare}), the model quickly maximizes the training accuracy. Providing less structure ({\sc Permute}) results in lower training accuracy when only a few training examples have been observed.  However, with additional training instances both sketches learn the correct behavior and generalize equally well.

% shows training and test accuracies for the two sketches discussed above when varying the number of training instances.  
%Here, training sequences are of length 3 and test sequences of length 8.
%As expected, providing less structure ({\sc Permute} sketch) results in a worse fit on the training set when given only few training examples.
%However, with more training examples the {\sc Permute} sketch with less prior structure generalize as well as the {\sc Compare} sketch. 
%For $256$ training examples both sketches learn the correct behavior and generalize to sequences that are over six times long.

\subsection{Addition}

Next, we applied \fnam{} to the problem of learning to add two n-digit numbers.
We rely on the standard elementary school addition algorithm, where the goal is to iterate over pairs of aligned digits, calculating the sum of each to yield the resulting sum.
The key complication arises when two digits sum to a two-digit number, requiring that the correct extra digit (a \emph{carry}) be carried over to the subsequent column.

We assume aligned pairs of digits as input, with a carry for the least significant digit (potentially $0$), and the length of the respective numbers.  
The sketches define the high-level operations through recursion, leaving the core addition to be learned from data. 

The specified high-level behaviour includes the recursive call template and the halting condition of the recursion (no remaining digits, line 2-3). The underspecified addition operation must take three digits from the previous call, the two digits to sum and a previous carry, and produce a single digit (the sum) and the resultant carry (lines 6a, 6b and 7a, 7b).  We introduce two sketches for inducing this behaviour: 

%, the {\sc Manipulate} (Listing~\ref{adder_manipulate}) and {\sc Choose} Listing~\ref{adder_choose}) sketches. Both sketches for the addition as input require the aligned pairs of digits, a carry for the least significant digit (potentially $0$), and the length of the respective numbers. %The sketches define the high-level operations through recursion, leaving the core addition to be learned from data. The specified high-level behavior includes the recursive call template and the halting condition of the recursion (no remaining digits, line 1-2). 
%The underspecified addition operation must take three digits from the previous call, the two digits to sum and a previous carry, and produce a single digit (the sum) and the resultant carry.

\bgroup
\createlinenumber{2}{}
\createlinenumber{3}{2}
\createlinenumber{4}{3}
\createlinenumber{5}{4}
\createlinenumber{6}{5}
\createlinenumber{7}{6a}
\createlinenumber{8}{}
\createlinenumber{9}{7a}
\createlinenumber{10}{6b}
\createlinenumber{11}{}
\createlinenumber{12}{7b}
\createlinenumber{13}{}
\createlinenumber{14}{}
\createlinenumber{15}{8}
\createlinenumber{16}{9}
\createlinenumber{17}{10}
\createlinenumber{18}{11}
\createlinenumber{19}{12}
\begin{figure}[]
\begin{lstlisting}[
  mathescape,
  caption={Manipulate sketch (a lines -- green) and the choose sketch (b lines -- blue) for Elementary Addition. Input data is used to fill data stack externally},
  label=adder_sketches,
  frame=single,
  basicstyle=\ttfamily\scriptsize,
  linebackgroundcolor={\ifnum\value{lstnumber}=7\color{mygreen}\else\ifnum\value{lstnumber}=8\color{mygreen}\else\ifnum\value{lstnumber}=9\color{mygreen}\else\ifnum\value{lstnumber}=10\color{myblue}\else\ifnum\value{lstnumber}=11\color{myblue}\else\ifnum\value{lstnumber}=12\color{myblue}\else\ifnum\value{lstnumber}=13\color{myblue}\else\ifnum\value{lstnumber}=14\color{myblue}\fi\fi\fi\fi\fi\fi\fi\fi}
]
: ADD-DIGITS 
  ( a1 b1...an bn carry n -- r1 r2...r_{n+1} ) 
  DUP 0 = IF
    DROP
  ELSE
    >R \ put n on R 
    { observe D0 D-1 D-2 -> tanh -> linear 70
      -> manipulate D-1 D-2 } 
    DROP
    { observe D0 D-1 D-2 -> tanh -> linear 10 
      -> choose 0 1 }
    { observe D-1 D-2 D-3 -> tanh -> linear 50
      -> choose 0 1 2 3 4 5 6 7 8 9 }
      >R SWAP DROP SWAP DROP SWAP DROP R>
    R> 1- SWAP >R  \ new_carry n-1 
    ADD-DIGITS \ call add-digits on n-1 subseq.
    R> \ put remembered results back on the stack
  THEN
;
\end{lstlisting}     
\end{figure}
\egroup

\begin{description}[leftmargin=0.5cm]

  \item {\sc Manipulate}. This sketch provides little prior procedural knowledge as it directly manipulates the \fnam{} machine state, filling in a carry and the result digits, based on the top three elements on the data stack (two digits and the carry).  Depicted in Listing~\ref{adder_sketches} in green.

  \item {\sc Choose}. Incorporating additional prior information, {\sc Choose} exactly specifies the results of the computation, namely the output of the first slot (line 6b) is the carry, and the output of the second one (line 7b) is the result digit, both conditioned on the two digits and the carry on the data stack. Depicted in Listing~\ref{adder_sketches} in blue.

\end{description}

%As underspecified content it leaves the manipulation of the two digits and a carry to produce a new digit (the sum) and the resulting carry, %and the calculation of the correct carry as behavior to be learned by the model.
The rest of the sketch code reduces the problem size by one and returns the solution by popping it from the return stack.

\begin{table}
\centering
\caption{Accuracy (Hamming distance) of Choose and Manipulate sketches in comparison to a Seq2Seq baseline on the addition problem. Note that lengths corresponds to the length of the input sequence (two times the number of digits of both numbers).}
\label{table:adder-results}
%\small
%\begin{tabular}{c|c|c|c || c|c|c}
\resizebox{\columnwidth}{!}{
\begin{tabular}{l rrr rrr}
\toprule
%\multicolumn{7}{c}{Addition}
& \multicolumn{3}{c}{Test Length 8} & \multicolumn{3}{c}{Test Length 64} \\
\cmidrule(l){2-4}
\cmidrule(l){5-7}
 Train Length:& 2 & 4 & 8 & 2 & 4 & 8 \\
 \midrule
 Seq2Seq  & 37.9 & 57.8 & 99.8 & 15.0 & 13.5 & 13.3 \\
 \fnam{} Choose & 100.0 & 100.0 & 100.0 & 100.0 & 100.0 & 100.0 \\
 \fnam{} Manipulate & 98.58  & 100.0 & 100.0 & 99.49 & 100.0 & 100.0\\
 \bottomrule
% Seq2Seq w/ Attention & \\
\end{tabular}
}
\end{table}

\paragraph{Quantitative Evaluation on Addition}

In a set of experiments analogous to those in our evaluation on Bubble sort, we demonstrate the performance of {\fnam} on the addition task by examining test set sequence lengths of 8 and 64 while varying the lengths of the training set instances (Table~\ref{table:adder-results}). The Seq2Seq model again fails to generalise to longer sequences than those observed during training. In comparison, both the {\sc Choose} sketch and the less structured {\sc Manipulate} sketch learn the correct sketch behaviour and generalise to all test sequence lengths (with an exception of {\sc Manipulate} which required more data to train perfectly). In additional experiments, we were able to successfully train both the {\sc Choose} and the {\sc Manipulate} sketches from sequences of input length $24$, and we tested them up to the sequence length of $128$, confirming their perfect training and generalisation capabilities. %Note that our experiments with {\sc Manipulate} include softmaxing values written on the stack. If this is removed, the sketch is able to learn only from the minimal sequence length of 2.

%We first tested the sketch on a training set of 200 single-digit addition examples. Both sketches successfully learned the addition, and generalised to sequences of length 8.
%Then we trained both sketches on a training set of 512 examples, with the length of the input sequence equal to $2$, $4$ and $8$ (see Table\ref{tab:adder-results}). The {\sc Choose} sketch generalised successfully to longer sequences (we tested it up to length of 64). The {\sc Manipulate} sketch both failed to train from longer sequences and failed to test on long sequences. We conjecture that this is due to difficulty of the direct state manipulation where the neural network in the slot needs to directly write the value on the data stack, and thus learn by itself the required output, as opposed to {\sc Choose} sketch where the user defines the output directly.

\subsection{Word Algebra Problems}

Word algebra problems (WAPs) are often used to assess the numerical reasoning abilities of schoolchildren.  Questions are short narratives which focus on numerical quantities, culminating with a question. For example:

\vspace{-1em}
\begin{quote}
\emph{A florist had 50 roses. If she sold 15 of them and then later picked 21 more, how many roses would she have?}
\end{quote}
\vspace{-1em}

Answering such questions requires both the understanding of language and of algebra --- one must know which numeric operations correspond to which phrase and how to execute these operations.

Previous work cast WAPs as a transduction task by mapping a question to a template of a mathematical formula, thus requiring manuall labelled formulas.
For instance, one formula that can be used to correctly answer the question in the example above is \code{(50 - 15) + 21 = 56}. In previous work, local classifiers~\citep{Roy2015SolvingGA, roy2015}, hand-crafted grammars~\citep{Kedziorski}, and recurrent neural models~\cite{bouchard-stenetorp-riedel:2016:EMNLP2016} have been used to perform this task.
Predicted formula templates may be marginalised during training~\citep{kushman-EtAl:2014:P14-1}, or evaluated directly to produce an answer.

In contrast to these approaches, {\fnam} is able to learn both, a soft mapping from text to algebraic operations \emph{and their execution}, without the need for manually labelled equations and no explicit symbolic representation of a formula.
%A {\fnam} sketch can be seen as defining the space of possible formula templates, which are then learned from input-output examples such that.

%While the methodology may vary with a largely hand-crafted grammar \citep{Kedziorski}
%, or with recurrent neural models \cite{bouchard-stenetorp-riedel:2016:EMNLP2016}, but in 

%constructs templates from many individual number and operator extraction classifiers \citep{roy2015},

%While the methodology may vary, in each approach notion of a formula template is explicit in the model.

 %\citep{kushman-EtAl:2014:P14-1}

\paragraph{Model description}

Our model is a fully end-to-end differentiable structure, consisting of a \fnam{} interpreter, a sketch, and a Bidirectional LSTM (BiLSTM) reader.

The BiLSTM reader reads the text of the problem and produces a vector representation (word vectors) for each word, concatenated from the forward and the backward pass of the BiLSTM network.
We use the resulting word vectors corresponding only to numbers in the text, numerical values of those numbers (encoded as one-hot vectors), and a vector representation of the whole problem (concatenation of the last and the first vector of the opposite passes) to initialise the \fnam{} heap $\heapb$.
This is done in an end-to-end fashion, enabling gradient propagation through the BiLSTM to the vector representations.
The process is depicted in Figure~\ref{model}.

The sketch, depicted in Listing~\ref{wap_partial} dictates the differentiable computation.\footnote{Due to space constraints, we present the core of the sketch here. For the full sketch, please refer to Listing \ref{wap_full_sketch} in the Appendix.}
First, it copies values from the heap $\heapb$ -- word vectors to the return stack $\rstackb$, and numbers (as one-hot vectors) on the data stack $\dstackb$.
Second, it contains four slots that define the space of all possible operations of four operators on three operands, all conditioned on the vector representations on the return stack.
These slots are i) permutation of the elements on the data stack, ii) choosing the first operator, iii) possibly swapping the intermediate result and the last operand, and iv) the choice of the second operator.
The final set of commands simply empties out the return stack $\rstackb$.
These slots define the space of possible operations, however, the model needs to learn how to utilise these operations in order to calculate the correct result.

% The model is trained with Adam~\citep{kingma2014adam}, with a learning rate of $0.02$. We randomly initialised word vectors of size $75$, with mean $0.0$ and stdev $0.1$. We employed gradient clipping to the norm of $1.0$.
\bgroup
\createlinenumber{1}{}
\createlinenumber{2}{1}
\createlinenumber{3}{2}
\createlinenumber{4}{3}
\createlinenumber{5}{4}
\createlinenumber{6}{}
\begin{figure}[]
\begin{lstlisting}[
  mathescape,
  caption=Core of the Word Algebra Problem sketch. The full sketch can be found in the Appendix.,
  label=wap_partial,
  frame=single,
  basicstyle=\ttfamily\scriptsize
]
\ first copy data from H: vectors to R and numbers to D
{ observe R0 R-1 R-2 R-3 -> permute D0 D-1 D-2 }
{ observe R0 R-1 R-2 R-3 -> choose + - * / }
{ observe R0 R-1 R-2 R-3 -> choose SWAP NOP }
{ observe R0 R-1 R-2 R-3 -> choose + - * / }
\ lastly, empty out the return stack
\end{lstlisting}  
\end{figure}
\egroup
\paragraph{Results}
We evaluate the model on the Common Core (CC) dataset, introduced by \citet{Roy2015SolvingGA}.
CC is notable for having the most diverse set of equation patterns, consisting of four operators (\code{+}, \code{-}, \code{$\times$}, \code{$\div$}), with up to three operands.

We compare against three baseline systems: (1) a local classifier with hand-crafted features \citep{Roy2015SolvingGA}, (2) a Seq2Seq baseline, and (3) the same model with a data generation component (GeNeRe)~\citet{bouchard-stenetorp-riedel:2016:EMNLP2016}.  All baselines are trained to predict the best equation, which is executed outside of the model to obtain the answer. In contrast, {\fnam} is trained end-to-end from input-output pairs and predicts the answer directly without the need for an intermediate symbolic representation of a formula.

Results are shown in Table~\ref{table:wap-results}. All RNN-based methods (bottom three) outperform the classifier-based approach.  
Our method slightly outperforms a Seq2Seq baseline, achieving the highest reported result on this dataset without data augmentation.

%local classifier-based approaches \citep{Roy2015SolvingGA}, approaching the maximum score obtainable for this task.  Our method slightly outperforms a Seq2Seq baseline

\begin{table}[]
\centering
\caption{Accuracies of models on the CC dataset. Asterisk denotes results obtained from \citet{bouchard-stenetorp-riedel:2016:EMNLP2016}. Note that GeNeRe makes use of additional data}% and is not directly comparable.}
\label{table:wap-results}
\begin{tabular}{lr}
\toprule
Model & Accuracy (\%) \\ 
\midrule
\multicolumn{2}{c}{\emph{Template Mapping}}\\
%\\ \hline
\citet{Roy2015SolvingGA} & 55.5 \\ 
Seq2Seq$^*$ \cite{bouchard-stenetorp-riedel:2016:EMNLP2016} & 95.0 \\ 
GeNeRe$^*$ \cite{bouchard-stenetorp-riedel:2016:EMNLP2016}   & 98.5 \\
\midrule
\multicolumn{2}{c}{\emph{Fully End-to-End}} \\ %\hline
\fnam{} & 96.0  \\ 
\bottomrule
\end{tabular}
\end{table}

%\subsubsection{Discussion}
%caveats
%- no loops (yet)
%- if we want a loop, we need to condition on one more element (otherwise it won't work)
%- not all formulas are semantically 'logical', but they work due to commutativity 
%advantages
%-end-to-end differentiable calculation!
%-no need for templating!
%-by looking at the permutation and ops we can reconstruct 'the formula'

\section{Discussion}
\label{sec:discussion}

{\fnam} bridges the gap between a traditional programming language and a modern machine learning architecture. However, as we have seen in our evaluation experiments, faithfully simulating the underlying abstract machine architecture introduces its own unique set of challenges. 

One such challenge is the additional complexity of performing even simple tasks when they are viewed in terms of operations on the underlying machine state. As illustrated in Table~\ref{table:sorter-results}, {\fnam} sketches can be effectively trained from small training sets (see Appendix~\ref{appendix:accuracy-examples}), and generalise perfectly to sequences of any length. However,  difficulty arises when training from sequences of modest lengths. Even when dealing with relatively short training length sequences, and with the program code optimisations employed, the underlying machine can unroll into a problematically large number states.
For problems whose machine execution is quadratic, like the sorting task (which at input sequences of length 4 has 120 machine states), we observe significant instabilities during training from backpropagating through such long RNN sequences, and consequent failures to train. In comparison, the addition problem was easier to train due to a comparatively shorter underlying execution RNNs.

The higher degree of prior knowledge provided played an important role in successful learning. For example, the {\sc Compare} sketch, which provides more structure, achieves higher accuracies when trained on longer sequences.
Similarly, employing softmax on the directly manipulated memory elements enabled perfect training for the {\sc Manipulate} sketch for addition.
Furthermore, it is encouraging to see that {\fnam} can be trained jointly with an upstream LSTM to provide strong procedural prior knowledge for solving a real-world NLP task.

%A second issue arises from incongruencies between how algorithmic data structures are typically used in a traditional language, and our newly introduced desire for them to learn behaviors which generalize to unseen data.  For instance, it is common in Forth implementations of sequence processing (Sec. \ref{sec:bubble-exec}) to include the sequence length on the input, and to store it during computation for use in the algorithm.  This is a sensible approach when working purely in the traditional programming paradigm, but in the context of learning it introduces information which influences the model's representations, prevents it from generalizing, and possibly leads to large memory requirements (when encoded with a one-hot vector).  This motivates investigation into which traditional language properties are most suitable to this new hybrid paradigm, and which representations can be used to circumvent the problem.

%When {\fnam} 
%in which the length of the sequence is appended to the input sequence, and passed repeatedly throughout the program.  This type of bookkeeping is not ideal in the machine learning context, where it influences the representations learned by the model.

%\code{BUBBLE}

\section{Related Work}
% \mat{include:terpret, adaptive compilation, autograd, neural compiler}

% \tim{and http://blob.lri.fr/publication/tcs.pdf}

%\tim{need to add Jacob Andreas' Neural Module Networks (NAACL paper)!}\mat{sure, it's related, someone add it!}
% \jason{For the CVPR introducing the idea}
%\Seb{each subpart should finish with a statement on how we differ, and maybe our advantages. }

\paragraph{Program Synthesis}

The idea of program synthesis is as old as Artificial Intelligence, and has a long history in computer science \citep{manna1971toward}. Whereas a large body of work has focused on using genetic programming \citep{koza1992genetic} to induce programs from  the given input-output specification \citep{nordin1997evolutionary}, there are also various Inductive Programming approaches \citep{kitzelmann2009inductive} aimed at inducing programs from incomplete specifications of the code to be implemented \citep{albarghouthi2013recursive, solar2006combinatorial}. We tackle the same problem of sketching, but in our case, we fill the sketches with neural networks able to learn the slot behaviour.

% Other approaches were taken too. Inductive Programming approaches \citep{kitzelmann2009inductive} .
% Heuristic search in special graph structures induced recursive programs that satisfy the input-output specifications . Lastly, SAT-based program synthesisers  successfully filled in simple underspecified bit-stream programmes.
% JN: 'programs' for consistency or is there a distinction?

% \paragraph{NLP} \tim{put this after Neural Approaches?} Stack-based abstract machines have been widely used in Natural Language Processing, particularly in the area of syntactic parsing~\cite{HY03b}, where they are often referred to as transition systems. These systems usually operate a buffer to read tokens from, and a single stack to push parse trees to. Our work is similar in spirit, in particular when considering recent neural transition systems~\cite{DBLP:journals/corr/DyerBLMS15}, but considers more general abstract machines that can be programmed, and taught, to do a wider range of tasks.

\paragraph{Probabilistic and Bayesian Programming}

Our work is closely related to probabilistic programming languages such as Church~\citep{goodman2012church}. 
They allow users to inject random choice primitives into programs as a way to define generative distributions over possible execution traces. In a sense, the random choice primitives in such languages correspond to the slots in our sketches. 
A core difference lies in the way we train the behaviour of slots: instead of calculating their posteriors using probabilistic inference, we estimate their parameters using backpropagation and gradient descent.
This is similar in-kind to \texttt{TerpreT}'s FMGD algorithm~\citep{gaunt2016terpret}, which is employed for code induction via backpropagation. In comparison, our model which optimises slots of neural networks surrounded by continuous approximations of code, enables the combination of procedural behaviour and neural networks.
In addition, the underlying programming and probabilistic paradigm in these programming languages is often functional and declarative, whereas our approach focuses on a procedural and discriminative view.      
By using an end-to-end differentiable architecture, it is easy to seamlessly connect our sketches to further neural input and output modules, such as an LSTM that feeds into the machine heap. %or a neural reinforcement learning agent that operates the neural machine.

\paragraph{Neural approaches}

Recently, there has been a surge of research in program synthesis, and execution in deep learning, with increasingly elaborate deep models. Many of these models were based on differentiable versions of abstract data structures~\citep{joulin2015inferring, grefenstette2015learning, kurach2015neural}, and a few abstract machines, such as the NTM~\citep{graves2014neural}, Differentiable Neural Computers~\citep{graves2016hybrid}, and Neural GPUs \citep{kaiser2015neural}. All these models are able to induce algorithmic behaviour from training data. Our work differs in that our differentiable abstract machine allows us to seemingly integrate code and neural networks, and train the neural networks specified by slots via backpropagation. % through code interpretation.
Related to our efforts is also the Autograd~\citep{maclaurin2015gradient}, which enables automatic gradient computation in pure Python code, but does not define nor use differentiable access to its underlying abstract machine.

% Subsequent work explored other memory structures, and controller structures to approach general computability.
% Neural stacks aided in learning binary addition and recognizing context-free languages , and were a crucial component in language transduction experiments, together with the neural queues, and deques . In other work, LSTM \citep{hochreiter1997long} controllers controlled connections between computational primitives to learn referencing and dereferencing pointers in various array and list access problems.

The work in neural approximations to abstract structures and machines naturally leads to more elaborate machinery able to induce and call code or code-like behaviour.
\citet{neelakantan2015neural} learned simple SQL-like behaviour--—querying tables from the natural language with simple arithmetic operations.
Although sharing similarities on a high level, the primary goal of our model is not induction of (fully expressive) code but its injection.
\cite{Andreas_2016_CVPR} learn to \emph{compose} neural modules to produce the desired behaviour for a visual QA task.
Neural Programmer-Interpreters~\citep{reed2015neural} learn to represent and execute programs, operating on different modes of an environment, and are able to incorporate decisions better captured in a neural network than in many lines of code (e.g. using an image as an input). Users inject prior procedural knowledge by training on program traces and hence require \emph{full} procedural knowledge. In contrast, we enable users to use their partial knowledge in sketches. 

Neural approaches to language compilation have also been researched, from compiling a language into neural networks \citep{siegelmann1994neural}, over building neural compilers~\citep{gruau1995neural} to adaptive compilation \citep{bunel2016adaptive}. However, that line of research did not perceive neural interpreters and compilers as a means of injecting procedural knowledge as we did.
% Generally previous neural approaches have been focusing on learning algorithms completely from scratch, and it is not obvious how to incorporate any structural priors. 
To the best of our knowledge, \fnam{} is the first working neural implementation of an abstract machine for an actual programming language, and this enables us to inject such priors in a straightforward manner.

% I toned down Turing-complete arguments, these are easy targets for reviewers to attack. 

\section{Conclusion and Future Work}
We have presented \fnam{}, a differentiable abstract machine for the Forth programming language, and showed how it can be used to complement programmers' prior knowledge through the learning of unspecified behaviour in Forth sketches.
%fill in missing behaviour in Forth program sketches. 
%Using only input-output pairs, \fnam{} is able to successfully learn sorting and addition algorithms.
The \fnam{} RNN successfully learns to sort and add, and solve word algebra problems, using only program sketches and program input-output pairs.
%We believe \fnam{} will be useful for addressing complex problems, such as machine reading and inference in knowledge bases where we could inject the recursive structure of theorem proving but leave unification open to be learned. \tim{quite a leap of faith}
% We believe \fnam{}, and the larger paradigm it helps establish, will be useful for addressing complex problems, such as machine reading and knowledge base inference, where higher-level reasoning is difficult to learn and may be easier to specify.
We believe \fnam{}, and the larger paradigm it helps establish, will be useful for addressing complex problems where low-level representations of the input are necessary, but higher-level reasoning is difficult to learn and potentially easier to specify.

%In future work we aim to explore the relationship between backpropagation and programming language semantics.  For instance, when performing a single conceptual operation whose implementation requires a series of simple operations on the machine state (e.g. Listing \ref{adder_choose}, line 7), one may wish to shortcut the flow of gradients around such low-entropy sequences using residual connections~\citep{highwaynets}.  An alternative solution to this problem would involve inducing hierarchies of actions, where a single more abstract action can be substituted for several repetitive actions, reducing the length of the unrolled action sequence. 

In future work, we plan to apply \fnam{} to other problems in the NLP domain, like machine reading and knowledge base inference. In the long-term, we see the integration of non-differentiable transitions (such as those arising when interacting with a real environment), as an exciting future direction which sits at the intersection of reinforcement learning and probabilistic programming. %Additionally, connecting \fnam{} with other differentiable models upstream and/or downstream is yet another direction for continued research.

%We also think it can be used as testbed for understanding how much prior structural bias a problem needs. 
%In the future we plan to focus on more user-friendly host languages, scaling up learning and execution further, and the integration of non-differentiable transitions (such as those arising when interacting with a real environment).  

% It did not just learn to make choices in a fixed recursion traces, it could also learn to operate the program counter and guarantee that recursions terminate. Using ideas from the programming language community, we could improve the runtime of the machine substantially. 

% Crucially, we not only learn to make choices in a fixed recursion trace, the \fnam{} can also learn to operate the program counter and guarantee that recursions terminate. 

% Future work:
% \begin{itemize}
% 	\item More practical languages (python abstract machine somewhat similar, scheme etc.)
% 	\item real world applications, sota somewhere
% 	\item scaling up
% 	\item memory management (garbage collection?)
% 	\item induction of complex programs
% 	\item curriculum and reusing of learned abstractions
% \end{itemize}

% gating, residual connections, 

\subsubsection*{Acknowledgments}

We thank Guillaume Bouchard, Danny Tarlow, Dirk Weissenborn, Johannes Welbl and the anonymous reviewers for fruitful discussions and helpful comments on previous drafts of this paper.
This work was supported by a Microsoft Research PhD Scholarship, an Allen Distinguished Investigator Award, and a Marie Curie Career Integration Award.

\bibliography{literature}

\begin{thebibliography}{36}
\providecommand{\natexlab}[1]{#1}
\providecommand{\url}[1]{\texttt{#1}}
\expandafter\ifx\csname urlstyle\endcsname\relax
  \providecommand{\doi}[1]{doi: #1}\else
  \providecommand{\doi}{doi: \begingroup \urlstyle{rm}\Url}\fi

\bibitem[Abadi et~al.(2015)Abadi, Agarwal, Barham, Brevdo, Chen, Citro,
  Corrado, Davis, Dean, Devin, Ghemawat, Goodfellow, Harp, Irving, Isard, Jia,
  Jozefowicz, Kaiser, Kudlur, Levenberg, Man\'{e}, Monga, Moore, Murray, Olah,
  Schuster, Shlens, Steiner, Sutskever, Talwar, Tucker, Vanhoucke, Vasudevan,
  Vi\'{e}gas, Vinyals, Warden, Wattenberg, Wicke, Yu, and
  Zheng]{tensorflow2015-whitepaper}
Abadi, Mart\'{\i}n, Agarwal, Ashish, Barham, Paul, Brevdo, Eugene, Chen,
  Zhifeng, Citro, Craig, Corrado, Greg~S., Davis, Andy, Dean, Jeffrey, Devin,
  Matthieu, Ghemawat, Sanjay, Goodfellow, Ian, Harp, Andrew, Irving, Geoffrey,
  Isard, Michael, Jia, Yangqing, Jozefowicz, Rafal, Kaiser, Lukasz, Kudlur,
  Manjunath, Levenberg, Josh, Man\'{e}, Dan, Monga, Rajat, Moore, Sherry,
  Murray, Derek, Olah, Chris, Schuster, Mike, Shlens, Jonathon, Steiner,
  Benoit, Sutskever, Ilya, Talwar, Kunal, Tucker, Paul, Vanhoucke, Vincent,
  Vasudevan, Vijay, Vi\'{e}gas, Fernanda, Vinyals, Oriol, Warden, Pete,
  Wattenberg, Martin, Wicke, Martin, Yu, Yuan, and Zheng, Xiaoqiang.
\newblock {TensorFlow}: Large-scale machine learning on heterogeneous systems,
  2015.
\newblock URL \url{http://tensorflow.org/}.
\newblock Software available from tensorflow.org.

\bibitem[Albarghouthi et~al.(2013)Albarghouthi, Gulwani, and
  Kincaid]{albarghouthi2013recursive}
Albarghouthi, Aws, Gulwani, Sumit, and Kincaid, Zachary.
\newblock Recursive program synthesis.
\newblock In \emph{Computer Aided Verification}, pp.\  934--950. Springer,
  2013.

\bibitem[Andreas et~al.(2016)Andreas, Rohrbach, Darrell, and
  Klein]{Andreas_2016_CVPR}
Andreas, Jacob, Rohrbach, Marcus, Darrell, Trevor, and Klein, Dan.
\newblock Neural module networks.
\newblock In \emph{Proceedings of IEEE Conference on Computer Vision and
  Pattern Recognition (CVPR)}, 2016.

\bibitem[ANSI(1994)]{ansforth1994}
ANSI.
\newblock \emph{{Programming Languages - Forth}}, 1994.
\newblock American National Standard for Information Systems, ANSI X3.215-1994.

\bibitem[Bouchard et~al.(2016)Bouchard, Stenetorp, and
  Riedel]{bouchard-stenetorp-riedel:2016:EMNLP2016}
Bouchard, Guillaume, Stenetorp, Pontus, and Riedel, Sebastian.
\newblock Learning to generate textual data.
\newblock In \emph{Proceedings of the Conference on Empirical Methods in
  Natural Language Processing (EMNLP)}, pp.\  1608--1616, 2016.

\bibitem[Brodie(1980)]{brodie1980starting}
Brodie, Leo.
\newblock \emph{{Starting Forth}}.
\newblock Forth Inc., 1980.

\bibitem[Bunel et~al.(2016)Bunel, Desmaison, Kohli, Torr, and
  Kumar]{bunel2016adaptive}
Bunel, Rudy, Desmaison, Alban, Kohli, Pushmeet, Torr, Philip~HS, and Kumar,
  M~Pawan.
\newblock Adaptive neural compilation.
\newblock In \emph{Proceedings of the Conference on Neural Information
  Processing Systems (NIPS)}, 2016.

\bibitem[Gaunt et~al.(2016)Gaunt, Brockschmidt, Singh, Kushman, Kohli, Taylor,
  and Tarlow]{gaunt2016terpret}
Gaunt, Alexander~L, Brockschmidt, Marc, Singh, Rishabh, Kushman, Nate, Kohli,
  Pushmeet, Taylor, Jonathan, and Tarlow, Daniel.
\newblock {TerpreT: A Probabilistic Programming Language for Program
  Induction}.
\newblock \emph{arXiv preprint arXiv:1608.04428}, 2016.

\bibitem[Goodman et~al.(2008)Goodman, Mansinghka, Roy, Bonawitz, and
  Tenenbaum]{goodman2012church}
Goodman, Noah, Mansinghka, Vikash, Roy, Daniel~M, Bonawitz, Keith, and
  Tenenbaum, Joshua~B.
\newblock Church: a language for generative models.
\newblock In \emph{Proceedings of the Conference in Uncertainty in Artificial
  Intelligence (UAI)}, pp.\  220--229, 2008.

\bibitem[Graves et~al.(2014)Graves, Wayne, and Danihelka]{graves2014neural}
Graves, Alex, Wayne, Greg, and Danihelka, Ivo.
\newblock {Neural Turing Machines}.
\newblock \emph{arXiv preprint arXiv:1410.5401}, 2014.

\bibitem[Graves et~al.(2016)Graves, Wayne, Reynolds, Harley, Danihelka,
  Grabska-Barwi{\'n}ska, Colmenarejo, Grefenstette, Ramalho, Agapiou,
  et~al.]{graves2016hybrid}
Graves, Alex, Wayne, Greg, Reynolds, Malcolm, Harley, Tim, Danihelka, Ivo,
  Grabska-Barwi{\'n}ska, Agnieszka, Colmenarejo, Sergio~G{\'o}mez,
  Grefenstette, Edward, Ramalho, Tiago, Agapiou, John, et~al.
\newblock Hybrid computing using a neural network with dynamic external memory.
\newblock \emph{Nature}, 538\penalty0 (7626):\penalty0 471--476, 2016.

\bibitem[Grefenstette et~al.(2015)Grefenstette, Hermann, Suleyman, and
  Blunsom]{grefenstette2015learning}
Grefenstette, Edward, Hermann, Karl~Moritz, Suleyman, Mustafa, and Blunsom,
  Phil.
\newblock {Learning to Transduce with Unbounded Memory}.
\newblock In \emph{Proceedings of the Conference on Neural Information
  Processing Systems (NIPS)}, pp.\  1819--1827, 2015.

\bibitem[Gruau et~al.(1995)Gruau, Ratajszczak, and Wiber]{gruau1995neural}
Gruau, Fr{\'e}d{\'e}ric, Ratajszczak, Jean-Yves, and Wiber, Gilles.
\newblock {A Neural compiler}.
\newblock \emph{Theoretical Computer Science}, 141\penalty0 (1):\penalty0
  1--52, 1995.

\bibitem[Hochreiter \& Schmidhuber(1997)Hochreiter and
  Schmidhuber]{hochreiter1997long}
Hochreiter, Sepp and Schmidhuber, J{\"u}rgen.
\newblock Long short-term memory.
\newblock \emph{Neural Computation}, 9\penalty0 (8):\penalty0 1735--1780, 1997.

\bibitem[Joulin \& Mikolov(2015)Joulin and Mikolov]{joulin2015inferring}
Joulin, Armand and Mikolov, Tomas.
\newblock {Inferring Algorithmic Patterns with Stack-Augmented Recurrent Nets}.
\newblock In \emph{Proceedings of the Conferences on Neural Information
  Processing Systems (NIPS)}, pp.\  190--198, 2015.

\bibitem[Kaiser \& Sutskever(2015)Kaiser and Sutskever]{kaiser2015neural}
Kaiser, {\L}ukasz and Sutskever, Ilya.
\newblock {Neural GPUs learn algorithms}.
\newblock In \emph{Proceedings of the International Conference on Learning
  Representations (ICLR)}, 2015.

\bibitem[King(1976)]{King:1976:SEP:360248.360252}
King, James~C.
\newblock {Symbolic Execution and Program Testing}.
\newblock \emph{Commun. ACM}, 19\penalty0 (7):\penalty0 385--394, 1976.

\bibitem[Kingma \& Ba(2015)Kingma and Ba]{kingma2014adam}
Kingma, Diederik and Ba, Jimmy.
\newblock {Adam: A Method for Stochastic Optimization}.
\newblock In \emph{Proceedings of the International Conference for Learning
  Representations (ICLR)}, 2015.

\bibitem[Kitzelmann(2009)]{kitzelmann2009inductive}
Kitzelmann, Emanuel.
\newblock {Inductive Programming: A Survey of Program Synthesis Techniques}.
\newblock In \emph{International Workshop on Approaches and Applications of
  Inductive Programming}, pp.\  50--73, 2009.

\bibitem[Koncel-Kedziorski et~al.(2015)Koncel-Kedziorski, Hajishirzi,
  Sabharwal, Etzioni, and Ang]{Kedziorski}
Koncel-Kedziorski, Rik, Hajishirzi, Hannaneh, Sabharwal, Ashish, Etzioni, Oren,
  and Ang, Siena.
\newblock {Parsing Algebraic Word Problems into Equations}.
\newblock \emph{Transactions of the Association for Computational Linguistics
  (TACL)}, 3:\penalty0 585--597, 2015.

\bibitem[Koza(1992)]{koza1992genetic}
Koza, John~R.
\newblock \emph{{Genetic Programming: On the Programming of Computers by Means
  of Natural Selection}}, volume~1.
\newblock MIT press, 1992.

\bibitem[Kurach et~al.(2016)Kurach, Andrychowicz, and
  Sutskever]{kurach2015neural}
Kurach, Karol, Andrychowicz, Marcin, and Sutskever, Ilya.
\newblock {Neural Random-Access Machines}.
\newblock In \emph{Proceedings of the International Conference on Learning
  Representations (ICLR)}, 2016.

\bibitem[Kushman et~al.(2014)Kushman, Artzi, Zettlemoyer, and
  Barzilay]{kushman-EtAl:2014:P14-1}
Kushman, Nate, Artzi, Yoav, Zettlemoyer, Luke, and Barzilay, Regina.
\newblock {Learning to Automatically Solve Algebra Word Problems}.
\newblock In \emph{Proceedings of the Annual Meeting of the Association for
  Computational Linguistics (ACL)}, pp.\  271--281, 2014.

\bibitem[Lau et~al.(2001)Lau, Wolfman, Domingos, and Weld]{lau2001learning}
Lau, Tessa, Wolfman, Steven~A., Domingos, Pedro, and Weld, Daniel~S.
\newblock Learning repetitive text-editing procedures with smartedit.
\newblock In \emph{Your Wish is My Command}, pp.\  209--226. Morgan Kaufmann
  Publishers Inc., 2001.

\bibitem[Maclaurin et~al.(2015)Maclaurin, Duvenaud, and
  Adams]{maclaurin2015gradient}
Maclaurin, Dougal, Duvenaud, David, and Adams, Ryan~P.
\newblock {Gradient-based Hyperparameter Optimization through Reversible
  Learning}.
\newblock In \emph{Proceedings of the International Conference on Machine
  Learning (ICML)}, 2015.

\bibitem[Manna \& Waldinger(1971)Manna and Waldinger]{manna1971toward}
Manna, Zohar and Waldinger, Richard~J.
\newblock Toward automatic program synthesis.
\newblock \emph{Communications of the ACM}, 14\penalty0 (3):\penalty0 151--165,
  1971.

\bibitem[Neelakantan et~al.(2015{\natexlab{a}})Neelakantan, Le, and
  Sutskever]{neelakantan2015neural}
Neelakantan, Arvind, Le, Quoc~V, and Sutskever, Ilya.
\newblock {Neural Programmer: Inducing latent programs with gradient descent}.
\newblock In \emph{Proceedings of the International Conference on Learning
  Representations (ICLR)}, 2015{\natexlab{a}}.

\bibitem[Neelakantan et~al.(2015{\natexlab{b}})Neelakantan, Vilnis, Le,
  Sutskever, Kaiser, Kurach, and Martens]{neelakantan2015adding}
Neelakantan, Arvind, Vilnis, Luke, Le, Quoc~V, Sutskever, Ilya, Kaiser, Lukasz,
  Kurach, Karol, and Martens, James.
\newblock {Adding Gradient Noise Improves Learning for Very Deep Networks}.
\newblock \emph{arXiv preprint arXiv:1511.06807}, 2015{\natexlab{b}}.

\bibitem[Nordin(1997)]{nordin1997evolutionary}
Nordin, Peter.
\newblock \emph{{Evolutionary Program Induction of Binary Machine Code and its
  Applications}}.
\newblock PhD thesis, der Universitat Dortmund am Fachereich Informatik, 1997.

\bibitem[Reed \& de~Freitas(2015)Reed and de~Freitas]{reed2015neural}
Reed, Scott and de~Freitas, Nando.
\newblock Neural programmer-interpreters.
\newblock In \emph{Proceedings of the International Conference on Learning
  Representations (ICLR)}, 2015.

\bibitem[Roy \& Roth(2015)Roy and Roth]{Roy2015SolvingGA}
Roy, Subhro and Roth, Dan.
\newblock {Solving General Arithmetic Word Problems}.
\newblock In \emph{Proceedings of the Conference on Empirical Methods in
  Natural Language Processing (EMNLP)}, pp.\  1743--1752, 2015.

\bibitem[Roy et~al.(2015)Roy, Vieira, and Roth]{roy2015}
Roy, Subhro, Vieira, Tim, and Roth, Dan.
\newblock Reasoning about quantities in natural language.
\newblock \emph{Transactions of the Association for Computational Linguistics
  (TACL)}, 3:\penalty0 1--13, 2015.

\bibitem[Siegelmann(1994)]{siegelmann1994neural}
Siegelmann, Hava~T.
\newblock {Neural Programming Language}.
\newblock In \emph{Proceedings of the Twelfth AAAI National Conference on
  Artificial Intelligence}, pp.\  877--882, 1994.

\bibitem[Solar-Lezama et~al.(2005)Solar-Lezama, Rabbah, Bod\'{\i}k, and
  Ebcio\u{g}lu]{solar2005programming}
Solar-Lezama, Armando, Rabbah, Rodric, Bod\'{\i}k, Rastislav, and Ebcio\u{g}lu,
  Kemal.
\newblock {Programming by Sketching for Bit-streaming Programs}.
\newblock In \emph{Proceedings of Programming Language Design and
  Implementation (PLDI)}, pp.\  281--294, 2005.

\bibitem[Solar-Lezama et~al.(2006)Solar-Lezama, Tancau, Bodik, Seshia, and
  Saraswat]{solar2006combinatorial}
Solar-Lezama, Armando, Tancau, Liviu, Bodik, Rastislav, Seshia, Sanjit, and
  Saraswat, Vijay.
\newblock {Combinatorial Sketching for Finite Programs}.
\newblock In \emph{ACM Sigplan Notices}, volume~41, pp.\  404--415, 2006.

\bibitem[Sutskever et~al.(2014)Sutskever, Vinyals, and Le]{Sutskever:2014}
Sutskever, Ilya, Vinyals, Oriol, and Le, Quoc~V.
\newblock {Sequence to Sequence Learning with Neural Networks}.
\newblock In \emph{Proceedings of the Conference on Neural Information
  Processing Systems (NIPS)}, pp.\  3104--3112, 2014.

\end{thebibliography}
\bibliographystyle{icml2017}

\clearpage

\onecolumn

\section*{Appendix}

\appendix

\section{Forth Words and their implementation}
\label{appendix:forth-words}
We implemented a small subset of available Forth words in \fnam{}. The table of these words, together with their descriptions is given in Table~\ref{appendix:forth_words}, and their implementation is given in Table~\ref{table:fnam_words}. The commands are roughly divided into 7 groups. These groups, line-separated in the table, are:
\begin{quote}
\begin{description}
    \item[Data stack operations] \code{\{num\}, 1+, 1-, DUP, SWAP, OVER, DROP, +, -, *, /}
    \item[Heap operations] \code{@, !}
    \item[Comparators] \code{>, <, =}
    \item[Return stack operations] \code{>R, R>, @R}
    \item[Control statements] \code{IF..ELSE..THEN, BEGIN..WHILE..REPEAT, DO..LOOP}
    \item[Subroutine control] \code{:, \{sub\}, ;, MACRO}
    \item[Variable creation] \code{VARIABLE, CREATE..ALLOT}
\end{description}
\end{quote}

\begin{table*}[!htb]
\caption{Forth words and their descriptions. TOS denotes top-of-stack, NOS denotes next-on-stack, DSTACK denotes the data stack, RSTACK denotes the return stack, and HEAP denotes the heap.}
\resizebox{\columnwidth}{!}{%
\label{appendix:forth_words}
\begin{tabular}{ p{0.3\textwidth}p{0.65\textwidth} } 
 \toprule
 Forth Word & Description\\
 \midrule
 \code{\{num\}} & Pushes \code{\{num\}} to DSTACK.\\
 \code{1+} & Increments DSTACK TOS by 1.\\
 \code{1-} & Decrements DSTACK TOS by 1.\\
 \code{DUP} & Duplicates DSTACK TOS.\\
 \code{SWAP} & Swaps TOS and NOS.\\
 \code{OVER} & Copies NOS and pushes it on the TOS.\\
 \code{DROP} & Pops the TOS (non-destructive).\\ 
 \code{+, -, *, /} & Consumes DSTACK NOS and TOS. Returns NOS \code{operator} TOS.\\
 \midrule
 \code{@} & Fetches the HEAP value from the DSTACK TOS address.\\
 \code{!} & Stores DSTACK NOS to the DSTACK TOS address on the HEAP.\\ 
 \midrule
 \code{>, <, =} & Consumes DSTACK NOS and TOS.\newline
 Returns 1 (TRUE) if NOS $> | < | =$ TOS respectivelly, 0 (FALSE) otherwise.\\
 \midrule
 \code{>R} & Pushes DSTACK TOS to RSTACK TOS, removes it from DSTACK.\\
 \code{R>} & Pushes RSTACK TOS to DSTACK TOS, removes it from RSTACK.\\
 \code{@R} & Copies the RSTACK TOS TO DSTACK TOS.\\ 
 \midrule
 \code{IF..ELSE..THEN} & Consumes DSTACK TOS, if it equals to a non-zero number (TRUE), executes commands between IF and ELSE. Otherwise executes commands between ELSE and THEN.\\
 \code{BEGIN..WHILE..REPEAT} &  Continually executes commands between WHILE and REPEAT while the code between BEGIN and WHILE evaluates to a non-zero number (TRUE).\\ %\hline
 \code{DO..LOOP} & Consumes NOS and TOS, assumes NOS as a limit, and TOS as a current index. Increases index by 1 until equal to NOS. At every increment, executes commands between DO and LOOP.\\
 \midrule
 \code{:} & Denotes the subroutine, followed by a word defining it.\\
 \code{\{sub\}} & Subroutine invocation, puts the program counter PC on RSTACK, sets PC to the subroutine address.\\
 \code{;} & Subroutine exit. Consumest TOS from the RSTACK and sets the PC to it.\\
 \code{MACRO} & Treats the subroutine as a macro function.\\
 \midrule
 \code{VARIABLE} & Creates a variable with a fixed address. Invoking the variable name returns its address.\\
 \code{CREATE..ALLOT} & Creates a variable with a fixed address. Do not allocate the next N addresses to any other variable (effectively reserve that portion of heap to the variable)\\
 \bottomrule
\end{tabular}
}
\end{table*}

\label{appendix:fnam-words}
\begin{table*}[!h]
\caption{Implementation of Forth words described in Table~\ref{appendix:forth_words}. Note that the variable creation words are implemented as fixed address allocation, and \code{MACRO} words are implemented with inlining.}
\label{table:fnam_words}
\centering
\resizebox{0.93\columnwidth}{!}{%
\begin{tabular}{ p{0.4\textwidth}p{0.55\textwidth} } 

    \toprule[1pt]
    % \hline
    % \rowcolor{gray!50}
    \textbf{Symbol} & \textbf{Explanation} \\
    % \hline
    \midrule[0.7pt]

    $\stacks$ & Stack, $\stacks \in \{ \dstack, \rstack \}$ \\

    $\buffer$ & Memory buffer, $\buffer \in \{ \dstackb, \rstackb, \heapb \}$ \\
    
    $\mathbf{p}$ & Pointer, $\mathbf{p} \in \{ \dstackt, \rstackt, \pc \}$ \\
    
    $\mathbf{R^{1\pm}}$ & Increment and decrement matrices (circular shift) \newline 
    $\mathbf{R}_{ij}^{1\pm} = \left\{\begin{matrix}
        1 &i \pm 1 \equiv j (mod \;   n))\\ 
        0 & otherwise
    \end{matrix}\right.$
    \\
    $\mathbf{R^{+}}, \mathbf{R^{-}}, \mathbf{R^{*}}, \mathbf{R^{/}}$ & Circular arithmetic operation tensors\newline 
    $\mathbf{R_{ijk}^{\{op\}}} = \left\{\begin{matrix}
        1 &i \{op\} j \equiv k (mod \;   n))\\ 
        0 & otherwise
    \end{matrix}\right.$
    \\
 
    \midrule[0.7pt]
    % \rowcolor{gray!50}
    \textbf{Pointer and value manipulation} & \textbf{Expression} \\
    \midrule[0.7pt]
    
    Increment $\pointer$ (or value $\x$) & $inc(\pointer) = \pointer^{T} \mathbf{R^{1+}} $ \\
    
    Decrement $\pointer$ (or value $\x$) & $dec(\pointer) = \pointer^{T} \mathbf{R^{1-}} $ \\
    
    Algebraic operation application & $\{op\}(\mathbf{a}, \mathbf{b}) = \mathbf{a}^T \mathbf{R^{\{op\}}} \mathbf{b}$ \\
    
    Conditional jump $\pointer$ & 
    $jump(\pc, \mathbf{a}): p = (pop_{\dstackb}() \code{=} TRUE)$ \newline
    $\pc \leftarrow p \pc + (1-p) a$ \\
    
    $\pointer^{-1}$ & Next on stack, $\pointer \leftarrow \pointer^{T} \mathbf{R^{1-}}$ \\

    \midrule[0.7pt]
    % \rowcolor{gray!50}
    \textbf{Buffer manipulation} &  \\
    \midrule[0.7pt]
    
    READ from $\buffer$ & $read_{\buffer}(\pointer) = \pointer^{T} \buffer$\\
    
    WRITE to $\buffer$ & $write_{\buffer}(\x, \pointer): \buffer \leftarrow \buffer - 
    \pointer \otimes \mathbf{1} \cdot \buffer + \x \otimes \pointer$ \\
    
    PUSH $\x$ onto $\stacks$ & $push_{\buffer}(\x):$
        $write_{\buffer}(\x, \pointer)$
        \quad [side-effect: $\dstackt \leftarrow inc(\dstackt)$] \\
    
    POP an element from $\stacks$ & $pop_{\buffer}() = read_{\buffer}(\pointer)$ \quad [side-effect: $\dstackt \leftarrow dec(\dstackt)$]  \\

    \midrule[0.7pt]
    % \rowcolor{gray!50}
    \textbf{Forth Word} & \\
    \midrule[0.7pt]
 
    Literal $\x$ & $push_{\dstackb}(\x)$ \\
    
    \code{1+} & $write_{\dstackb}(inc(read_{\dstackb}(\dstackt)), \dstackt)$ \\
    
    \code{1-} & $write_{\dstackb}(dec(read_{\dstackb}(\dstackt)), \dstackt)$ \\
    
    \code{DUP} & $push_{\dstackb}(read_{\dstackb}(\dstackt))$ \\
    
    \code{SWAP} &  $x = read_{\dstackb}(\dstackt)$, $y = read_{\dstackb}(\dstackn)$ \newline
                   :$write_{\dstackb}(\dstackt, y)$ , $write_{\dstackb}(\dstackn, x)$ \\
 
    \code{OVER} & $push_{\dstackb}(read_{\dstackb}(\dstackt))$ \\
    
    \code{DROP} & $pop_{\dstackb}()$ \\
    
    \code{+, -, *, /} & $write_{\dstackb}(\{op\}(read_{\dstackb}(\dstackn), read_{\dstackb}(\dstackt)), \dstackt)$ \\
    \midrule
    
    \code{@} & $read_{\heapb}(\dstackt)$ \\
    \code{!} & $write_{\heapb}(\dstackt, \dstackn)$ \\
    
    \midrule
    
    \code{<} & \code{SWAP >} \\
    
    \code{>} &  
    $e_1 = \sum_{i=0}^{n-1} i * \dstackt_{i}$, $e_2 = \sum_{i=0}^{n-1} i * \dstackn_{i}$ \newline
    $p = \phi_{pwl}(\mathbf{e_1} - \mathbf{e_2})$, where $\phi_{pwl}(x) = min(max(0, x + 0.5), 1)$\newline
    $p  \textbf{1} + (p - 1)  \textbf{0}$ \\
    
    \code{=} & $p = \phi_{pwl}(\dstackt, \dstackn)$ \newline
    $p \textbf{1} + (p - 1)  \textbf{0}$ \\
    
    \midrule
    
    \code{>R} & $push_{\rstackb}(\dstackt)$ \\
    
    \code{R>} & $pop_{\rstackb}()$ \\
    
    \code{@R} & $write_{\dstackb}(\dstackt, read_{\rstackb}(\rstackt))$ \\
    
    \midrule
    
    \code{IF..\textsubscript{1}ELSE..\textsubscript{2}THEN} & 
     $p = (pop_{\dstackb}() \code{=} \textbf{0})$ \newline
     $p * .._{1} + (1 - p) * .._{2}$
    \\
 
    \code{BEGIN..\textsubscript{1}WHILE..\textsubscript{2}REPEAT} & 
        $.._{1}$
    $jump(c, .._{2})$
    \\

    \code{DO..LOOP} & 
    
    $start = \pc, current = inc(pop_{\dstackb}()), limit = pop_{\dstackb}()$\newline
    $p = (current = limit)$\newline
    $jump(p, ..), jump(c, start)$\\

    \bottomrule[1pt]
\end{tabular}
}
\end{table*}

\clearpage
\section{Bubble sort algorithm description}
\label{appendix:bubblesort}

An example of a Forth program that implements the Bubble sort algorithm is shown in Listing \ref{bubblesort}. We provide a description of how the first iteration of this algorithm is executed by the Forth abstract machine:

The program begins at line 11, putting the sequence \stack{2 4 2 7} on the data stack $D$, followed by the sequence length \code{4}.\footnote{Note that Forth uses Reverse Polish Notation and that the top of the data stack is 4 in this example.} It then calls the \code{SORT} word.

\begin{tabular}{p{0.3cm} | p{2cm} | p{3.5cm} | p{1cm} | p{5cm}}
 & $D$ & $R$ & $c$ & comment \\ \hline
 1 & [] & [] & 11 & execution start \\
 2 & [2 4 2 7 4] & [] & 8 & pushing sequence to $D$, calling SORT subroutine puts A\textsubscript{SORT} to $R$ \\
\end{tabular}

For a sequence of length $4$, \code{SORT} performs a do-loop in line 9 that calls the \code{BUBBLE} sub-routine $3$ times. It does so by decrementing the top of $D$ with the \code{1-} word to $3$. Subsequently, $3$ is duplicated on $D$ by using \code{DUP}, and $0$ is pushed onto $D$.

\begin{tabular}{p{0.3cm} | p{2cm} | p{3.5cm} | p{1cm} | p{5cm}}
3 & [2 4 2 7 3] & [A\textsubscript{SORT}] & 9 & 1- \\ 
4 & [2 4 2 7 3 3] & [A\textsubscript{SORT}] & 9 & DUP\\
6 & [2 4 2 7 3 3 0] & [A\textsubscript{SORT}] & 9 & 0\\
\end{tabular}

\code{DO} consumes the top two stack elements $3$ and $0$ as the limit and starting point of the loop, leaving the stack $D$ to be \stack{2,4,2,7,3}. 
We use the return stack  $R$ as a temporary variable buffer and push $3$ onto it using the word \code{>R}. This drops $3$ from $D$, which we copy from $R$ with \code{R@}

\begin{tabular}{p{0.3cm} | p{2cm} | p{3.5cm} | p{1cm} | p{5cm}}
7 & [2 4 2 7 3] & [Addr\textsubscript{SORT}] & 9 & DO \\
8 & [2 4 2 7] & [Addr\textsubscript{SORT} 3] & 9 & \textgreater R\\ 
9 & [2 4 2 7 3] & [Addr\textsubscript{SORT} 3] & 9 & @R \\
\end{tabular}

Next, we call \code{BUBBLE} to perform one iteration of the bubble pass, (calling \code{BUBBLE} $3$ times internally), and consuming $3$. 
Notice that this call puts the current program counter onto $R$, to be used for the program counter $c$ when exiting \code{BUBBLE}. 

Inside the \code{BUBBLE} subroutine, \code{DUP} duplicates $3$ on $R$. \code{IF} consumes the duplicated $3$ and interprets is as TRUE. \code{>R} puts $3$ on $R$.

\begin{tabular}{p{0.3cm} | p{2cm} | p{3.5cm} | p{1cm} | p{5cm}}
10 & [2 4 2 7 3] & [A\textsubscript{SORT} 3 A\textsubscript{BUBBLE}] & 0 & calling BUBBLE subroutine puts A\textsubscript{BUBBLE} to $R$\\
11 & [2 4 2 7 3 3] & [A\textsubscript{SORT} 3 A\textsubscript{BUBBLE}] & 1 & DUP\\
12 & [2 4 2 7 3] & [A\textsubscript{SORT} 3 A\textsubscript{BUBBLE}] & 1 & IF\\
13 & [2 4 2 7] & [A\textsubscript{SORT} 3 A\textsubscript{BUBBLE} 3] & 1 & \textgreater R\\
\end{tabular}

Calling \code{OVER} twice duplicates the top two elements of the stack, to test them with \code{<}, which tests whether $2<7$.
\code{IF} tests if the result is TRUE ($0$), which it is, so it executes \code{SWAP}.

\begin{tabular}{p{0.3cm} | p{2cm} | p{3.5cm} | p{1cm} | p{5cm}}
14 & [2 4 2 7 2 7] & [A\textsubscript{SORT} 3 A\textsubscript{BUBBLE} 3] & 2 & OVER OVER\\
15 & [2 4 2 7 1] & [A\textsubscript{SORT} 3 A\textsubscript{BUBBLE} 3] & 2 & \textless \\
16 & [2 4 2 7] & [A\textsubscript{SORT} 3 A\textsubscript{BUBBLE} 3] & 2 & IF\\
17 & [2 4 7 2] & [A\textsubscript{SORT} 3 A\textsubscript{BUBBLE} 3] & 2 & SWAP\\
\end{tabular}

To prepare for the next call to \code{BUBBLE} we move $3$ back from the return stack $R$ to the data stack $D$ via \code{R>}, \code{SWAP} it with the next element, put it back to $R$ with \code{>R}, decrease the TOS with \code{1-} and invoke \code{BUBBLE} again. Notice that $R$ will accumulate the analysed part of the sequence, which will be recursively taken back.

\begin{tabular}{p{0.3cm} | p{2cm} | p{3.5cm} | p{1cm} | p{5cm}}
18 & [2 4 7 2 3] & [A\textsubscript{SORT} 3 A\textsubscript{BUBBLE}] & 3 & R \textgreater\\
19 & [2 4 7 3 2] & [A\textsubscript{SORT} 3 A\textsubscript{BUBBLE}] & 3 & SWAP\\
20 & [2 4 7 3] & [A\textsubscript{SORT} 3 A\textsubscript{BUBBLE} 2] & 3 & \textgreater R\\
21 & [2 4 7 2] & [A\textsubscript{SORT} 3 A\textsubscript{BUBBLE} 2] & 3 & 1-\\
22 & [2 4 7 2] & [A\textsubscript{SORT} 3 A\textsubscript{BUBBLE} 2] & 0 & ...BUBBLE\\
\end{tabular}

When we reach the loop limit we drop the length of the sequence and exit \code{SORT} using the \code{;} word, which takes the return address from $R$. At the final point, the stack should contain the ordered sequence \stack{7 4 2 2}.

% \clearpage
% \subsection*{B. Forth Adder}
% \begin{lstlisting}[
%   mathescape,
%   caption=Full code for adding two digits and a carry. The code is long due to the fact that the adding of two numbers in Differentiable Forth is currently realised with multiple executions of 1+. ,
%   label=highschooladd,
%   frame=single,
%   basicstyle=\ttfamily\scriptsize
% ]
% : DIGIT+
%     DUP 1 = IF >R
%         DUP 9 = IF
%             DROP 0 R> ELSE 1+ R> 1-
%         THEN
%     THEN
%     >R
%       BEGIN DUP WHILE
%         1- SWAP DUP 9 = IF
%           R> 1+ >R
%           DROP 0
%         ELSE
%           1+ SWAP
%         THEN
%       REPEAT
%       DROP
%       R>
%     ;
% \end{lstlisting}

\twocolumn

\section{Learning and Run Time Efficiency}

\subsection{Accuracy per training examples}
\label{appendix:accuracy-examples}
\paragraph{Sorter}

When measuring the performance of the model as the number of training \emph{instances} varies, we can observe the benefit of additional prior knowledge to the optimisation process.  We find that when stronger prior knowledge is provided ({\sc Compare}), the model quickly maximises the training accuracy. Providing less structure ({\sc Permute}) results in lower testing accuracy initially, however, both sketches learn the correct behaviour and generalise equally well after seeing $256$ training instances. Additionally, it is worth noting that the {\sc Permute} sketch was not always able to converge into a result of the correct length, and both sketches are not trivial to train.

In comparison, Seq2Seq baseline is able to generalise only to the sequence it was trained on (Seq2Seq trained and tested on sequence length 3). When training it on sequence length 3, and testing it on a much longer sequence length of 8, Seq2Seq baseline is not able to achieve more than $45\%$ accuracy.

\begin{figure}[h!]
     \centering
     \includegraphics[width=0.5\textwidth]{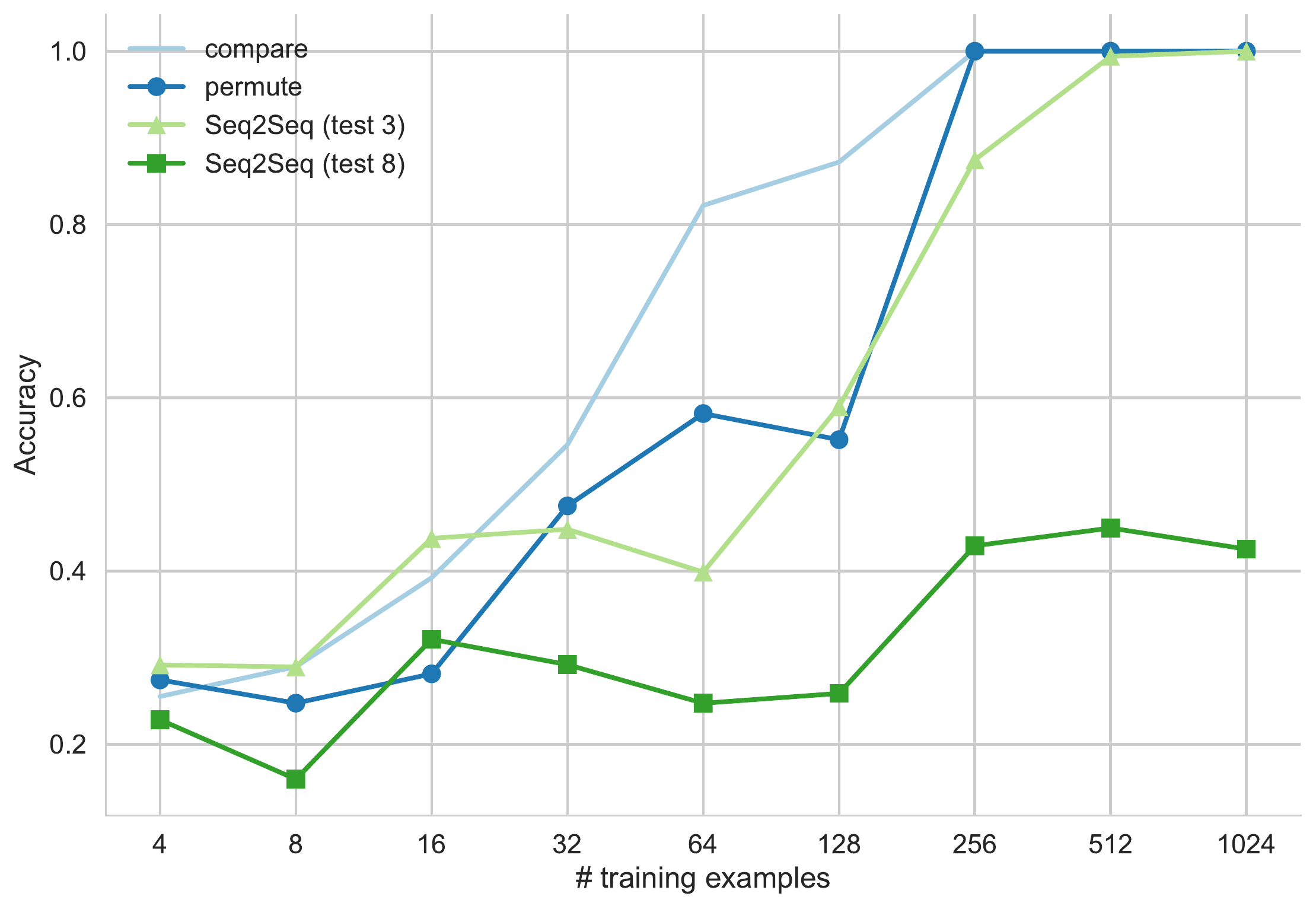}
     %\subfloat[][Varying training sequence lengths.]{\includegraphics[width=0.33\textwidth]{results_train}\label{fig:res2}}\\
     %\subfloat[][Train and test accuracy for varying test sequence lengths.]{\includegraphics[width=0.32\textwidth]{results_test}\label{fig:res3}}
     %\hfill
     \caption{Accuracy of models for varying number of training examples, trained on input sequence of length 3 for the Bubble sort task. Compare, permute, and Seq2Seq (test 8) were tested on sequence lengths 8, and Seq2Seq (test 3) was tested on sequence length 3. }
     \label{appendix:sorter-examples}
\end{figure}

\begin{figure}[h!]
     \centering
     \includegraphics[width=0.5\textwidth]{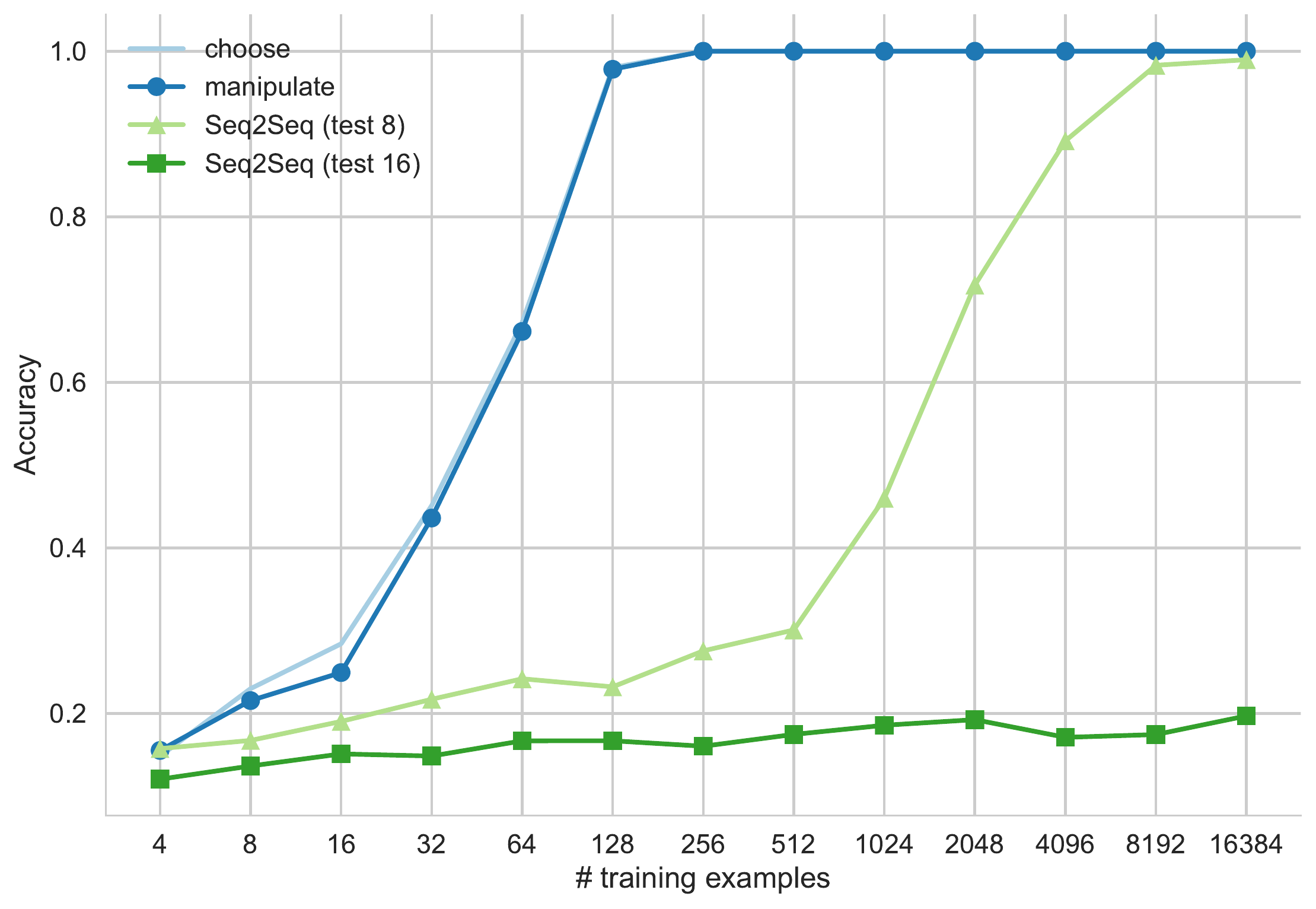}
     %\subfloat[][Varying training sequence lengths.]{\includegraphics[width=0.33\textwidth]{results_train}\label{fig:res2}}\\
     %\subfloat[][Train and test accuracy for varying test sequence lengths.]{\includegraphics[width=0.32\textwidth]{results_test}\label{fig:res3}}
     %\hfill
     \caption{Accuracy of models for varying number of training examples, trained on input sequence of length 8 for the addition task. Manipulate, choose, and Seq2Seq (test 16) were tested on sequence lengths 16, and Seq2Seq (test 8) was tested on sequence length 8.}
     \label{appendix:adder-examples}
\end{figure}

\paragraph{Adder} We tested the models to train on datasets of increasing size on the addition task. The results, depicted in Table~\ref{appendix:adder-examples} show that both the choose and the manipulate sketch are able to perfectly generalise from $256$ examples, trained on sequence lengths of $8$, tested on $16$. In comparison, the Seq2Seq baseline achieves $98\%$ when trained on $16384$ examples, but only when tested on the input of the same length, $8$. If we test Seq2Seq as we tested the sketches, it is unable to achieve more $19.7\%$.

\subsection{Program Code Optimisations}

\begin{figure}[h!]
     \centering
     \includegraphics[width=0.5\textwidth]{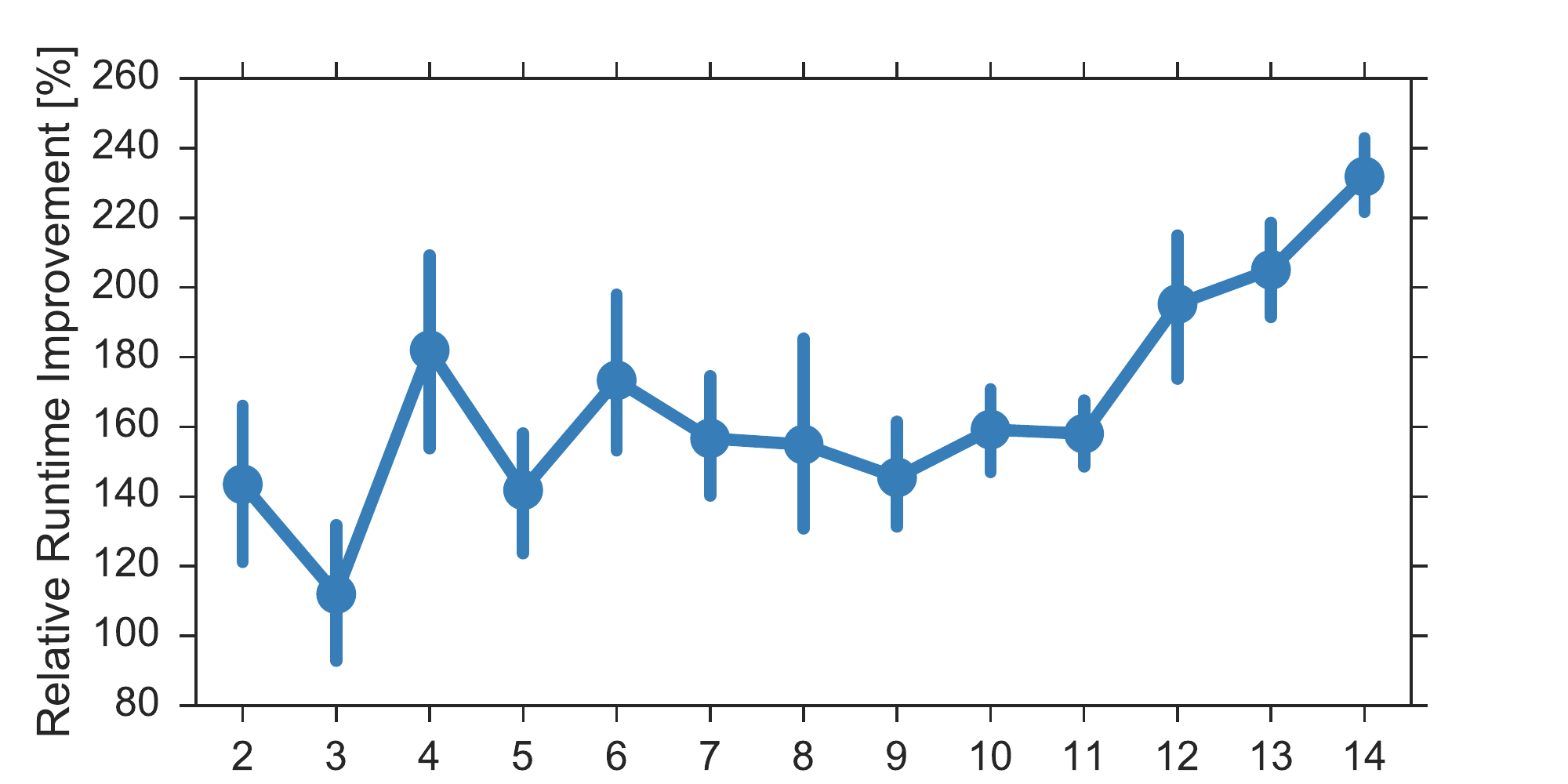}
     \caption{Relative speed improvements of program code optimisations for different input sequence lengths (\emph{bottom}).}
     \label{results}
\end{figure}

We measure the runtime of Bubble sort on sequences of varying length with and without the optimisations described in Section \ref{sec:optim}. 
The results of ten repeated runs are shown in Figure \ref{results} and demonstrate large relative improvements for symbolic execution and interpolation of if-branches compared to non-optimised \fnam{} code.

\section{\fnam{} execution of a Bubble sort sketch}
\label{appendix:bubble-execution}

Listing~\ref{bubblesort} (lines 3b and 4b -- in blue) defines the \code{BUBBLE} word as a sketch capturing several types of prior knowledge. In this section, we describe the {\sc Permute} sketch. In it, we assume \code{BUBBLE} involves a recursive call, that terminates at length 1, and that the next \code{BUBBLE} call takes as input some function of the current length and the top two stack elements.

The input to this sketch are the sequence to be sorted and its length decremented by one, $n-1$ (line 1). These inputs are expected on the data stack. After the length ($n-1$) is duplicated for further use with \code{DUP}, the machine tests whether it is non-zero (using \code{IF}, which consumes the TOS during the check). If $n-1>0$, it is stored on the $R$ stack for future use (line 2).

At this point (line 3b) the programmer only knows that a decision must be made based on the top two data stack elements \code{D0} and \code{D-1} (comparison elements), and the top return stack, \code{R0} (length decremented by 1). Here the precise nature of this decision is unknown but is limited to variants of permutation of these elements, the output of which produce the input state to the decrement \code{-1} and the recursive \code{BUBBLE} call (line 4b). At the culmination of the call, \code{R0}, the output of the learned slot behaviour, is moved onto the data stack using \code{R>}, and execution proceeds to the next step.

%the top two stack elements \code{D0} and \code{D-1}, as well as the top return stack element \code{R0}, must be observed and based on this observation, permuted to produce the input state to the recursive \code{BUBBLE} call (line 2 in Listing \ref{bubblesketch}). 
%After the call we put whatever currently is on the return stack (as manipulated by the slot) onto the data stack using \code{R>}, and then move on with the next step.

Figure \ref{rnn} illustrates how portions of this sketch are executed on the \fnam{} RNN. The program counter initially resides at \code{>R} (line 3 in $\mathbf{P}$), as indicated by the vector $\pc$, next to program $\mathbf{P}$. Both data and return stacks are partially filled ($\rstack$ has 1 element, $\dstack$ has 4), and we show the content both through horizontal one-hot vectors and their corresponding integer values (colour coded). The vectors $\mathbf{d}$ and  $\mathbf{r}$ point to the top of both stacks, and are in a one-hot state as well. In this execution trace, the slot at line 4 is already showing optimal behaviour: it remembers the element on the return stack (4) is larger and executes \code{BUBBLE} on the remaining sequence with the counter $n$ subtracted by one, to 1.

\section{Experimental details}
\label{appendix:experimental-details}

The parameters of each sketch are trained using Adam~\citep{kingma2014adam}, with gradient clipping (set to 1.0) and gradient noise~\citep{neelakantan2015adding}. 
We tuned the learning rate, batch size, and the parameters of the gradient noise in a random search on a development variant of each task.

\subsection{Seq2Seq baseline}

The Seq2Seq baseline models are single-layer networks with LSTM cells of $50$ dimensions.

The training procedure for these models consists of 500 epochs of Adam optimisation, with a batch size of 128, a learning rate of 0.01, and gradient clipping when the L2 norm of the model parameters exceeded 5.0.  We vary the size of training and test data (Fig.~\ref{appendix:sorter-examples}), but observe no indication of the models failing to reach convergence under these training conditions.

\subsection{Sorting}

The Permute and Compare sketches in Table~\ref{table:sorter-results} were trained on a randomly generated train, development and test set containing 256, 32 and 32 instances, respectively. Note that the low number of dev and test instances was due to the computational complexity of the sketch.

The batch size was set to a value between $64$ and $16$, depending on the problem size, and we used an initial learning rate of $1.0$.

\subsection{Addition}

We trained the addition Choose and Manipulate sketches presented in Table~\ref{table:adder-results} on a randomly generated train, development and test sets of sizes 512, 256, and 1024 respectively. The batch size was set to $16$, and we used an initial learning rate of $0.05$

\subsection{Word Algebra Problem}

The Common Core (CC) dataset~\citep{Roy2015SolvingGA} is partitioned into a train, dev, and test set containing 300, 100, and 200 questions, respectively. The batch size was set to $50$, and we used an initial learning rate of $0.02$. 
The BiLSTM word vectors were initialised randomly to vectors of length $75$. The stack width was set to $150$ and the stack size to $5$.

% \clearpage
% \onecolumn

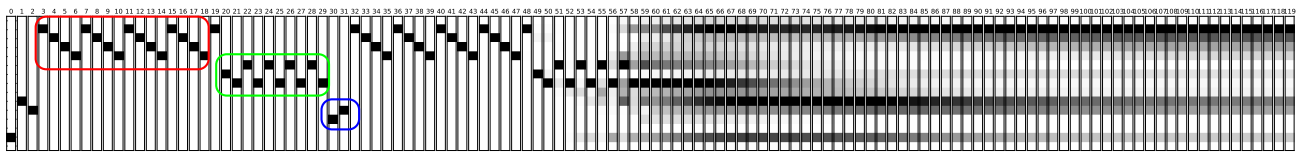
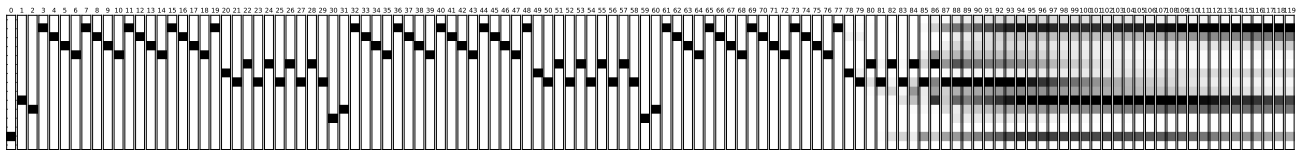
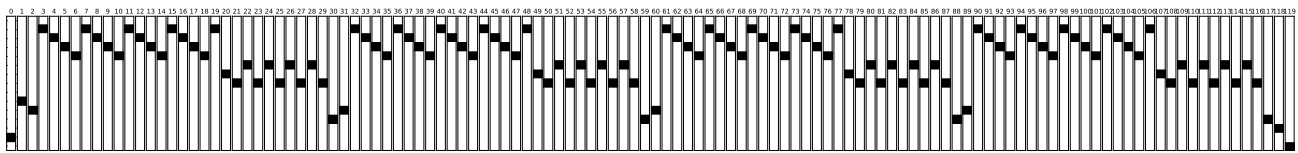
\begin{figure*}
     \centering
     \subfloat[][Program Counter trace in early stages of training.]{\begin{tikzpicture}
\node[anchor=south west,inner sep=0] at (0,0) {\includegraphics[width=1.0\textwidth]{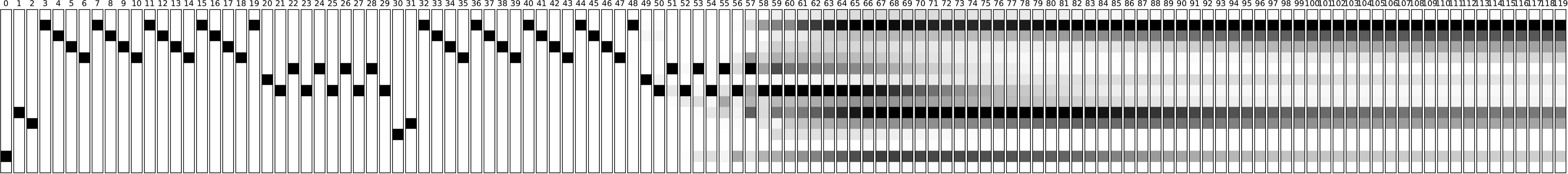}};
\draw[red,thick,rounded corners] (0.4,1.1) rectangle (2.7,1.8);
\draw[green,thick,rounded corners] (2.8,0.75) rectangle (4.3,1.3);
\draw[blue,thick,rounded corners] (4.2,0.3) rectangle (4.7,0.7);
\end{tikzpicture}}\\

     \subfloat[][Program Counter trace in the middle of training.]{\includegraphics[width=1.0\textwidth]{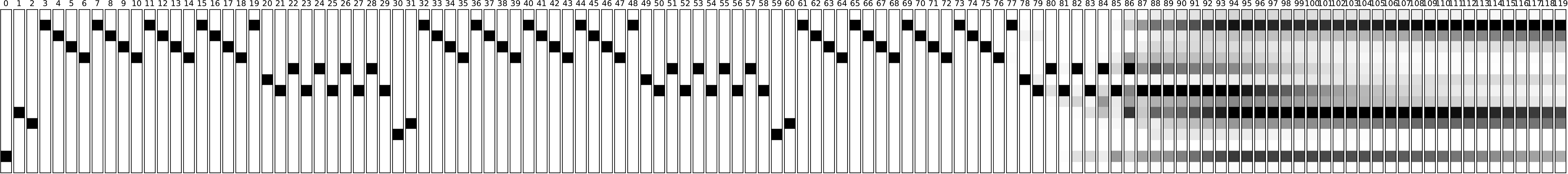}\label{fig:res2}}\\
     \subfloat[][Program Counter trace at the end of training.]{\includegraphics[width=1.0\textwidth]{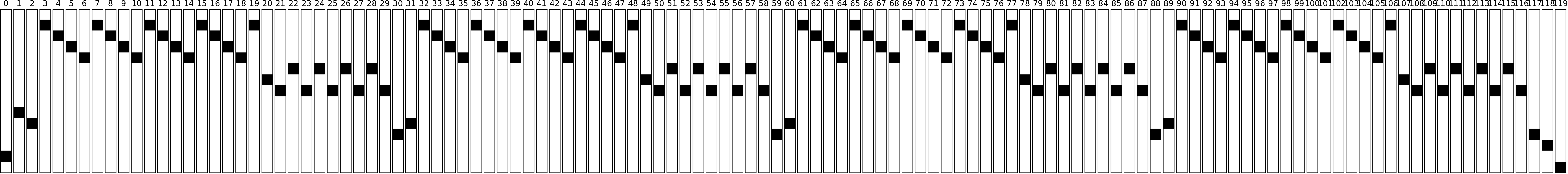}\label{fig:res3}}\\
     \caption{Program Counter traces for a single example at different stages of training BubbleSort in Listing~\ref{bubblesort} (red: successive recursion calls to \code{BUBBLE}, green: successive returns from the recursion, and blue: calls to SORT). The last element in the last row is the halting command, which only gets executed after learning the correct slot behaviour.}
     \label{steady_state}

\end{figure*}

\section{Qualitative Analysis on BubbleSort of PC traces}
\label{appendix:}

In Figure~\ref{steady_state} we visualise the program counter traces. The trace follows a single example from start, to middle, and the end of the training process. In the beginning of training, the program counter starts to deviate from the one-hot representation in the first $20$ steps (not observed in the figure due to unobservable changes), and after two iterations of \code{SORT}, \fnam{} fails to correctly determine the next word. After a few training epochs \fnam{} learns better permutations which enable the algorithm to take crisp decisions and halt in the correct state.

\clearpage
\onecolumn
\section{The complete Word Algebra Problem sketch}

The Word Algebra Problem (WAP) sketch described in Listing~\ref{wap_partial} is the core of the model that we use for WAP problems.
However, there were additional words before and after the core which took care of copying the data from the heap to data and return stacks, and finally emptying out the return stack.

The full WAP sketch is given in Listing~\ref{wap_full_sketch}. We define a \code{QUESTION} variable which will denote the address of the question vector on the heap. Lines 4 and 5 create \code{REPR\_BUFFER} and \code{NUM\_BUFFER} variables and denote that they will occupy four sequential memory slots on the heap, where we will store the representation vectors and numbers, respectively. Lines 7 and 8 create variables \code{REPR} and \code{NUM} which will denote addresses to current representations and numbers on the heap. Lines 10 and 11 store \code{REPR\_BUFFER} to \code{REPR} and \code{NUM\_BUFFER} to \code{NUM}, essentially setting the values of variables \code{REPR} and \code{NUM} to starting addresses allotted in lines 4 and 5. Lines 14-16 and 19-20 create macro functions \code{STEP\_NUM} and \code{STEP\_REPR} which increment the \code{NUM} and \code{REPR} values on call. These macro functions will be used to iterate through the heap space. Lines 24-25 define macro functions \code{CURRENT\_NUM} for fetching the current number, and \code{CURRENT\_REPR} for fetching representation values. Lines 28-32 essentially copy values of numbers from the heap to the data stack by using the  \code{CURRENT\_NUM} and \code{STEP\_NUM} macros. After that line 35 pushes the question vector, and lines 36-40 push the word representations of numbers on the return stack.

Following that, we define the core operations of the sketch. Line 43 permutes the elements on the data stack (numbers) as a function of the elements on the return stack (vector representations of the question and numbers). Line 45 chooses an operator to execute over the TOS and NOS elements of the data stack (again, conditioned on elements on the return stack). Line 47 executes a possible swap of the two elements on the data stack (the intermediate result and the last operand) conditioned on the return stack. Finally, line 49 chooses the last operator to execute on the data stack, conditioned on the return stack.

The sketch ends with lines 52-55 which empty out the return stack.

\begin{figure}
%  \begin{minipage}{0.45\textwidth}

\begin{lstlisting}[
  mathescape,
  caption=The complete Word Algebra Problem sketch,
  label=wap_full_sketch,
  frame=single,
  basicstyle=\ttfamily\scriptsize
]
\ address of the question on H
VARIABLE QUESTION
\ allotting H for representations and numbers
CREATE REPR_BUFFER 4 ALLOT
CREATE NUM_BUFFER 4 ALLOT
\ addresses of the first representation and number
VARIABLE REPR
VARIABLE NUM

REPR_BUFFER REPR !
NUM_BUFFER NUM !

\ macro function for incrementing the pointer to numbers in H
MACRO: STEP_NUM
  NUM @ 1+ NUM !
;

\ macro function for incrementing the pointer to representations in H
MACRO: STEP_REPR
  REPR @ 1+ REPR !
;

\ macro functions for fetching current numbers and representations
MACRO: CURRENT_NUM NUM @ @ ;
MACRO: CURRENT_REPR REPR @ @ ;

\ copy numbers to D
CURRENT_NUM
STEP_NUM
CURRENT_NUM
STEP_NUM
CURRENT_NUM

\ copy question vector, and representations of numbers to R
QUESTION @ >R
CURRENT_REPR >R
STEP_REPR
CURRENT_REPR >R
STEP_REPR
CURRENT_REPR >R

\ permute stack elements, based on the question and number representations
{ observe R0 R-1 R-2 R-3 -> permute D0 D-1 D-2 }
\ choose the first operation
{ observe R0 R-1 R-2 R-3 -> choose + - * / }
\ choose whether to swap intermediate result and the bottom number 
{ observe R0 R-1 R-2 R-3 -> choose SWAP NOP }
\ choose the second operation
{ observe R0 R-1 R-2 R-3 -> choose + - * / }

\ empty out R
R> DROP
R> DROP
R> DROP
R> DROP
\end{lstlisting}
\end{figure}

\end{document}